\documentclass[sigconf]{acmart}
\AtBeginDocument{%
  }

\setcopyright{cc}
\copyrightyear{2026}
\acmYear{2026}
\setcctype{by}
\acmDOI{10.1145/3770855.3818914}
\acmConference[KDD 2026] {Proceedings of the 32nd ACM SIGKDD Conference on Knowledge Discovery and Data Mining V.2}{August 9--13, 2026}{Jeju Island, Republic of Korea.}
\acmBooktitle{Proceedings of the 32nd ACM SIGKDD Conference on Knowledge Discovery and Data Mining V.2 (KDD 2026), August 9--13, 2026, Jeju Island, Republic of Korea}
\acmISBN{979-8-4007-2259-2/2026/08}

\usepackage{booktabs}
\usepackage{multirow}
\usepackage[table]{xcolor}
\usepackage{enumitem}
\usepackage{graphicx}
\usepackage{tabularx}

\usepackage{listings}
\usepackage[most]{tcolorbox} %
\tcbuselibrary{listings, breakable, skins} %

\lstdefinelanguage{json}{
    basicstyle=\ttfamily\scriptsize,
    numbers=left,
    numberstyle=\scriptsize,
    stepnumber=1,
    numbersep=8pt,
    showstringspaces=false,
    breaklines=true,
    frame=none,
    backgroundcolor=\color{gray!5},
    literate=
     *{0}{{{\color{blue}0}}}{1}
      {1}{{{\color{blue}1}}}{1}
      {2}{{{\color{blue}2}}}{1}
      {3}{{{\color{blue}3}}}{1}
      {4}{{{\color{blue}4}}}{1}
      {5}{{{\color{blue}5}}}{1}
      {6}{{{\color{blue}6}}}{1}
      {7}{{{\color{blue}7}}}{1}
      {8}{{{\color{blue}8}}}{1}
      {9}{{{\color{blue}9}}}{1}
      {:}{{{\color{red}{:}}}}{1}
      {,}{{{\color{red}{,}}}}{1}
      {\{}{{{\color{black}{\{}}}}{1}
      {\}}{{{\color{black}{\}}}}}{1}
      {[}{{{\color{black}{[}}}}{1}
      {]}{{{\color{black}{]}}}}{1},
}

\newtcblisting{promptbox}[1][]{
  colback=gray!5!white,
  colframe=gray!75!black,
  title=\textbf{#1},
  fonttitle=\bfseries,
  enhanced,              %
  listing only,
  listing options={
    language=json,       %
    basicstyle=\normalsize\rmfamily, %
    breaklines=true,     %
    breakatwhitespace=true,
    columns=fullflexible,%
    keepspaces=true,
    showstringspaces=false
  }
}

\begin{document}

\title{SpaCellAgent: A Self-Evolving LLM-Based Multi-Agent Framework for Trajectory Analysis}

\author{Songhan Wang}
\email{243352479@st.usst.edu.cn}
\authornote{Equal contribution.}
\orcid{0009-0008-9467-2284}
\affiliation{%
  \institution{University of Shanghai for Science and Technology}
  \city{Shanghai}
  \country{China}
}

\author{Haoang Chi}
\email{haoangchi618@gmail.com}
\authornote{Corresponding author.}
\affiliation{%
  \institution{National University of Defense Technology}
  \city{Changsha}
  \country{China}
}

\author{He Li}
\email{lihemaster117@gmail.com}
\affiliation{%
  \institution{Hong Kong Baptist University}
  \city{Kowloon}
  \country{Hong Kong}
}
\affiliation{%
  \institution{National University of Defense Technology}
  \city{Changsha}
  \country{China}
}

\author{Zhiheng Zhang}
\email{zhangzhiheng@mail.shufe.edu.cn}
\affiliation{%
    \institution{Shanghai University of Finance and Economics}
  \city{Shanghai}
  \country{China}
}

\author{Jiayan Yuan}
\email{242482685@st.usst.edu.cn}
\authornotemark[1]
\affiliation{%
  \institution{University of Shanghai for Science and Technology}
  \city{Shanghai}
  \country{China}
}

\author{Cheems Wang}
\email{cheemswang@mail.tsinghua.edu.cn}
\affiliation{%
  \institution{Tsinghua University}
  \city{Beijing}
  \country{China}
}

\author{Hao Peng}
\email{penghao@buaa.edu.cn}
\affiliation{%
  \institution{Beihang University}
  \city{Beijing}
  \country{China}
}

\author{Xinwang Liu}
\email{xinwangliu@nudt.edu.cn}
\affiliation{%
  \institution{National University of Defense Technology}
  \city{Changsha}
  \country{China}
}

\author{Wenjing Yang}
\email{wenjing.yang@nudt.edu.cn}
\authornotemark[2]
\affiliation{%
  \institution{National University of Defense Technology}
  \city{Changsha}
  \country{China}
}

\renewcommand{\shortauthors}{Songhan Wang et al.}

\begin{abstract}
Spatial and Single-cell transcriptomics are transformative in deciphering cellular dynamics. As the fundamental paradigm for reconstructing cell developmental paths, trajectory inference (TI) is critical. However, existing methods require extensive manual intervention and proficiency in heterogeneous tools, posing a significant barrier to efficient TI analysis. To bridge this gap, we propose SpaCellAgent, an autonomous large language model (LLM) multi-agent framework that automates end-to-end spatiotemporal analysis and narrative generation. SpaCellAgent utilizes a multi-agent architecture for strategic workflow planning, a dynamic tool-orchestration engine for adaptive algorithm selection, and a self-evolution module that iteratively refines performance through feedback. We evaluate SpaCellAgent on six heterogeneous datasets encompassing complex temporal developmental trajectories, diverse sequencing platforms, and spatially-resolved tissue architectures. SpaCellAgent consistently demonstrates over 40\% improvement in analytical efficiency while maintaining expert-aligned performance. By converting natural language specifications into optimized analytical workflows and fully automating the pipeline, SpaCellAgent democratizes advanced spatiotemporal modeling and establishes a scalable, agent-driven paradigm for computational biology. The code and materials are available at \url{https://github.com/LittleXH-shw/SpaCellAgent}.

\end{abstract}



\maketitle

\section{Introduction}
\begin{figure*}[t]
    \centering
    \includegraphics[width=1\linewidth]{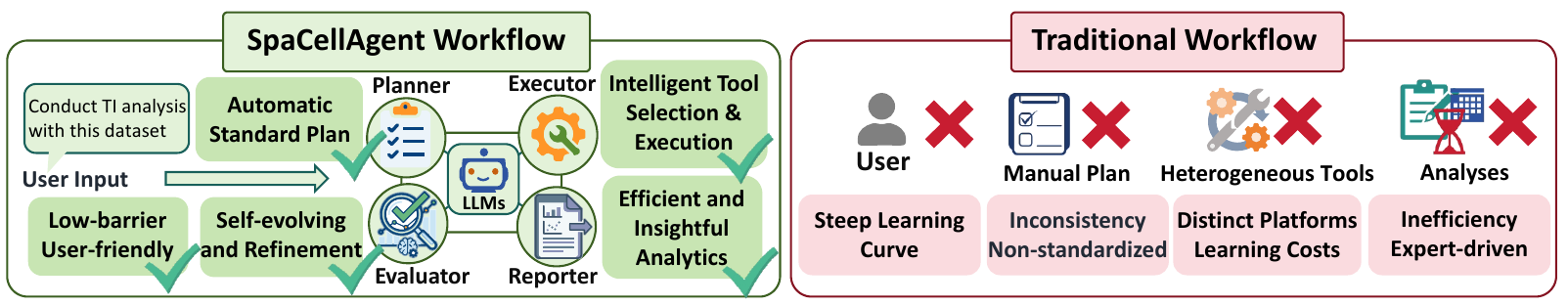}
    \caption{Comparison between SpaCellAgent and the traditional workflow. Our framework utilizes an LLM-driven multi-agent system to autonomously translate natural language queries and spatial data into biological insights. By orchestrating code generation and execution, it replaces complex, non-standardized traditional workflows with a streamlined, automated pipeline.}
    \Description{}
    \label{fig:1}

\end{figure*}
Single-cell RNA sequencing (scRNA-seq) has revolutionized our understanding of cellular diversity and developmental processes by enabling transcriptome profiling at individual cell resolution \cite{tanay2017scaling}. This technology captures the heterogeneity within cell populations \cite{tang2009mrna}, revealing rare cell types \cite{shalek2013single}, transitional states \cite{trapnell2014dynamics}, and dynamic expression patterns that were previously obscured in bulk measurements \cite{macosko2015highly}. Recently, the rapid development of spatial transcriptomics technologies has enabled the preservation of gene expression with spatial coordinates, allowing the mapping of cellular states within their native tissue architectures \cite{rodriques2019slide,chen2022spatiotemporal}. These approaches provide unprecedented insights into how cellular states evolve both temporally and spatially \cite{scialdone2016resolving}, offering a comprehensive view of developmental trajectories \cite{papalexi2018single}, disease mechanisms \cite{moffitt2018molecular}, and tissue homeostasis \cite{rao2021exploring,schaum2018single}. However, systematically uncovering molecular mechanisms from such high-dimensional cellular states necessitates intricate analysis pipelines \cite{hu2021spagcn}. Within this methodological landscape, Trajectory Inference (TI) stands out as a pivotal instrument for investigating cellular development and inferring cell fate, effectively reconstructing the continuous progression of biological processes from static snapshots \cite{cannoodt2016computational}.

To operationalize cellular development trajectory analysis, a diverse array of computational approaches has been established, including Monocle \cite{qiu2017reversed}, PAGA \cite{wolf2019paga}, Slingshot \cite{street2018slingshot}, and Diffusion Pseudotime (DPT) \cite{haghverdi2016diffusion}. These methods are grounded in distinct mathematical frameworks, ranging from graph-based algorithms to principal curve fitting and diffusion maps, to infer cellular ordering and branching structures. However, despite their theoretical advancements, these methods exhibit inconsistent performance across varying data dimensionalities and trajectory topologies \cite{saelens2019comparison}, thereby necessitating manual selection and heuristic tuning by domain experts. 

Recent advances in large language models (LLMs) have catalyzed a paradigm shift from manual tool-centric analysis to autonomous LLMs agent-driven exploration \cite{wang2023scientific, boiko2023autonomous}. Beyond code generation and complex reasoning \cite{guo2025deepseek,achiam2023gpt}, LLMs can now orchestrate heterogeneous computational tools and democratize scientific interpretation by bridging raw outputs with meaningful insights \cite{m2024augmenting,dai2024deep}. Despite this transformative potential, the integration of LLMs agent workflows into single-cell lineage reconstruction remains nascent, lacking an end-to-end closed-loop solution. Crucially, existing approaches lack automated mechanisms for self-evolution and iterative refinement; their static nature often yields suboptimal inference results and imposes a prohibitive manual burden. Consequently, there is an imperative for a unified framework that leverages the biological reasoning abilities of pre-trained LLMs as a cognitive core, autonomously orchestrating TI workflows.

To fill these gaps, we introduce SpaCellAgent, an autonomous LLM-driven framework that performs end-to-end spatial analysis and biological narrative generation. In contrast to traditional static and rigidly defined pipelines \cite{bacher2016design,wagner2020lineage}, SpaCellAgent introduces a collaborative multi-agent architecture that dynamically assigns specialized roles, such as planner, executor, evaluator, and reporter, to each LLM. Specifically, the planner analyzes trajectory analysis objectives and topological data properties to decompose the lineage reconstruction process into executable milestones, while the executor autonomously identifies optimal algorithmic strategies through a dynamic tool orchestration engine to generate robust implementation code. The evaluator assesses the quality of the reconstructed lineage, diagnosing specific TI anomalies such as implausible branching and pseudotime inversion, while synthesizing corrective feedback for iterative improvement through a self-refinement mechanism. Additionally, SpaCellAgent incorporates a self-evolution module that persistently accumulates reusable error fixes and validated analysis templates across tasks, enabling continuous improvement and knowledge accumulation. 
Extensive benchmarking demonstrates that SpaCellAgent outperforms state-of-the-art (SOTA) baselines in topological fidelity while reducing the analysis time by 41.2\%. Crucially, it reconciles scalability with robustness, delivering stable, expert-aligned trajectory analysis performance.

In general, the main contributions of this work can be summarized as follows:
\begin{itemize}[noitemsep, topsep=0pt, parsep=0pt, partopsep=0pt]
\item We propose SpaCellAgent, an autonomous multi-agent framework that performs end-to-end TI analysis by integrating scRNA-seq and spatial transcriptomics data, and generates interpretable biological reports derived from the TI results and other downstream analyses.
\item SpaCellAgent couples data-driven strategy selection with a self-evolving refinement mechanism, enabling the system to autonomously adapt to diverse biological heterogeneities while continuously accumulating knowledge to optimize analysis robustness.
\item Extensive experiments demonstrate SpaCellAgent's expert-aligned performance, and its 41.2\% improvement in analytical efficiency compared to manual expert workflows. Therefore, our SpaCellAgent drastically shortens the time-to-insight for complex trajectory analysis.
\end{itemize}

\section{Related Work}
\subsection{Trajectory Analysis in Single-Cell and Spatial Omics}
The computational reconstruction of cellular developmental trajectories has evolved through distinct methodological paradigms \cite{cannoodt2016computational, saelens2019comparison}. In the realm of scRNA-seq, foundational tools established the premise of ordering cells along a latent pseudotime \cite{haghverdi2016diffusion}. Monocle pioneered this approach using minimum spanning trees (MST) and reversed graph embedding to resolve branching lineages \cite{qiu2017reversed}. Slingshot introduced principal curves for stable multi-lineage inference \cite{street2018slingshot}, while PAGA utilized graph abstraction to effectively reconcile clustering with trajectory topology \cite{wolf2019paga}. 
The advent of spatial transcriptomics has catalyzed a paradigm shift in developmental modeling, necessitating the integration of physical topology into lineage reconstruction. Recent frameworks, such as SpaceFlow~\cite{ren2022identifying} and STORIES~\cite{huizing2025stories}, have extended TI by incorporating spatial neighborhood constraints to model tissue-scale gene expression gradients. Despite these methodological strides, the existing TI landscape remains fragmented and heavily reliant on manual expertise. Researchers are often constrained by heterogeneous computational ecosystems, necessitating experience-dependent method choice and inefficient manual exploration of the hyperparameter space. This lack of standardization not only hampers reproducibility but also creates a prohibitive barrier for domain scientists. To address these challenges, we propose SpaCellAgent, an LLM-driven agentic framework that autonomously orchestrates adaptive algorithm selection and iterative code refinement, ensuring robust and reproducible analysis across diverse biological contexts.

\subsection{LLM and Agent Systems for Scientific Discovery}

The advent of LLMs, such as Gemini 2.5~\cite{comanici2025gemini}, has led to a paradigm shift in code generation, natural language reasoning, and complex task decomposition. Building on these capabilities, general-purpose multi-agent architectures, such as ChatDev \cite{qian2024chatdev}, have demonstrated how specialized agents can collaborate to autonomously execute intricate software engineering tasks. Recently, this agent-driven paradigm has notably extended to physics, where frameworks like ChemCrow \cite{m2024augmenting} have successfully demonstrated the efficacy of LLMs in automating material property prediction and orchestrating chemical synthesis. However, in biomedicine and bioinformatics, the adoption of such workflows has been predominantly limited to textual tasks, such as literature retrieval and hypothesis formulation \cite{mathur2025pyevocell,xiao2024cellagent}. The potential for driving closed-loop, empirical data analysis, particularly for multi-step computational biology workflows, remains largely underexplored. 
Existing LLM agent frameworks, primarily optimized for general-purpose software engineering or text processing, lack the domain-specific abilities for biological data analysis. Specifically, they do not support the autonomous orchestration of heterogeneous TI analysis algorithms, nor do they possess the error-recovery mechanisms necessary for processing high-dimensional, noisy omics data. To date, a unified framework capable of executing end-to-end, closed-loop trajectory analysis tasks remains absent. Our SpaCellAgent fills this gap by establishing an LLM-driven autonomous framework that dynamically optimizes algorithmic selection, delivering end-to-end TI analysis through a robust, self-optimizing workflow.

\section{Preliminary}
\begin{figure*}[t]
    \centering
    \includegraphics[width=1\linewidth]{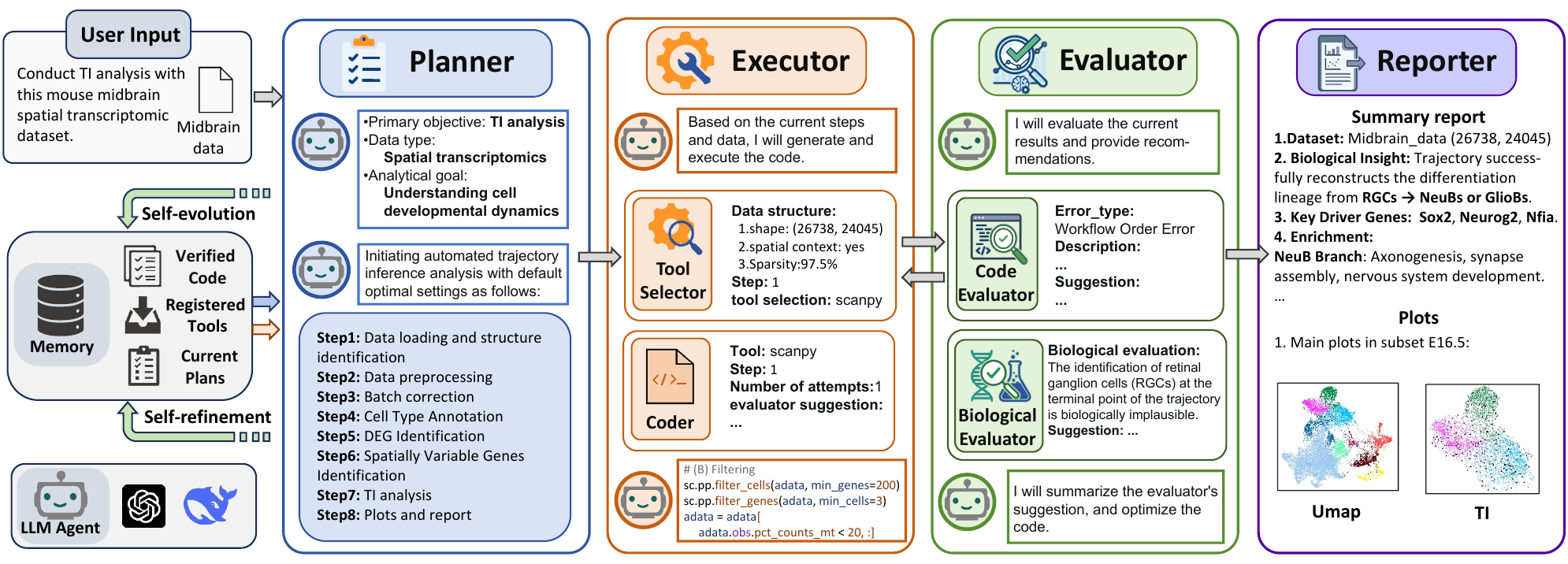}
    \caption{The framework of SpaCellAgent. The planner decomposes user queries into executable steps for the executor, while the evaluator enforces dual verification for syntactic and biological validity. The reporter then synthesizes the final biological insights. Arrows are color-coded: blue (input for planner), orange (input for executor), and green (evaluator's feedback for self-refinement and self-evolution). Validated plans are archived in dual-layer memory to drive self-evolution. }
    \Description{}
    \label{fig:2}
\end{figure*}

\label{sec:preliminary}


\subsection{Trajectory Inference Analysis in Cellular Data}
\label{subsec:ti-prelim}
Formally, let a single-cell dataset be denoted as ${D} = \{ \mathbf{X}, \mathbf{S} \}$. The gene expression profile is represented by a matrix $\mathbf{X} \in \mathbb{R}^{N \times G}$, where $N$ denotes the number of cells and $G$ represents the number of genes. Each row $\mathbf{x}_i \in \mathbb{R}^G$ corresponds to the high-dimensional expression vector of cell $i$.
In the context of spatial transcriptomics, each cell is additionally associated with a spatial coordinate vector $\mathbf{s}_i \in \mathbb{R}^D$ (typically $D=2$ for tissue sections), forming a spatial coordinate matrix $\mathbf{S} \in \mathbb{R}^{N \times D}$.
A common and advantageous approach is to model cellular relationships as a graph $G = (V, E)$, with nodes $V$ representing cells. The adjacency matrix $\mathbf{A} \in {0, 1}^{N \times N}$ is constructed based on pairwise transcriptomic similarity for scRNA-seq data. For spatial transcriptomics, it is built by integrating the expression matrix with physical coordinates, enabling the capture of the local manifold structure.
The primary objective of TI is to reconstruct a dynamic developmental process from these static snapshots. Mathematically, this can be formulated as learning a mapping $f: (\mathbf{X}, \mathbf{S}) \to \mathcal{T}$,
that projects the high-dimensional data onto a low-dimensional latent manifold $\mathcal{T}$, which represents the continuous space of all possible cell states along the process. Crucially, the inference of $\mathcal{T}$ is operationally realized by jointly determining:
\begin{enumerate}
    \item \textbf{Pseudotime Ordering ($\boldsymbol{\tau}$):} A continuous scalar value $\tau_i \in [0, 1]$ that assigns each cell a coordinate on the manifold $\mathcal{T}$, representing its progression level. for each cell $i$, representing its progression level along the biological process, where $\tau=0$ denotes the root state and $\tau=1$ denotes the terminal state.
    \item \textbf{Lineage Structure:} A connectivity graph or a set of principal curves describing the global topology (e.g., linear, bifurcating, or tree-structured) of cell fate decisions.
\end{enumerate}
In spatial-aware TI, the inferred pseudotime $\boldsymbol{\tau}$ must be smooth with respect to both the transcriptomic manifold $\mathbf{X}$ and the spatial graph defined by $\mathbf{S}$. 

\subsection{LLM-driven Autonomous Agents}
LLM-driven autonomous agents represent a paradigm shift in which LLMs serve as the ``brain'' for planning, reasoning, and action execution. While the individual agent demonstrates impressive capabilities in task decomposition, it is prone to instability in complex scenarios. Consequently, recent research focuses on multi-agent collaboration, which mimics human group dynamics by assigning distinct roles to specialized agents. By facilitating inter-agent interaction and iterative feedback, this collaborative architecture overcomes the limitations of a single agent, offering superior robustness and accuracy for handling sophisticated workflows such as computational biology tasks.

\section{Method}

\subsection{Framework Overview}
In this section, we introduce the overview of the SpaCellAgent. As shown in Figure \ref{fig:2}, SpaCellAgent is an autonomous, LLM-driven framework designed to organize complex trajectory analysis on single-cell and spatial transcriptomics data. Distinct from static code generation tools, our frameworks implement a dynamic, closed-loop workflow powered by multi-agent collaboration. Upon receiving a natural language task description, the system initiates a collaborative process involving distinct agent roles. This process encompasses data perception, where input characteristics are inspected; strategic planning, where optimal TI and downstream analysis methods are selected; and autonomous execution, where executable Python or R scripts are generated and verified. The workflow culminates in the generation of a comprehensive report, integrating visual outputs with biological interpretation, thereby streamlining the path from raw data to scientific hypothesis formulation.
\vspace{-4pt}
\subsection{Agent Roles and Functionalities}

\textbf{Planner agent}.
The planner agent acts as the system's reasoning engine, decomposing high-level user intents into executable logics. It accepts a dual input: the natural language query and a structured metadata profile of the target dataset, encompassing gene expression counts, sparsity patterns, and spatial modalities. Leveraging an LLM with specialized planning instructions, the agent synthesizes a directed action sequence encoded in a JSON file. 
Each node in this sequence is assigned a unique ID and a descriptive label, establishing a machine-readable backbone that facilitates robust downstream execution and human-in-the-loop verification.
\textbf{Executor agent}.
The executor agent bridges the gap between abstract planning and concrete execution. It consists of two tightly coupled components responsible for dynamic tool retrieval and context-aware code generation.

\textit{Tool selector}. This component functions as a semantic router, mapping high-level task descriptions to specific computational tools. It queries a curated registry containing both standard single-cell methods and spatial-aware algorithms. By utilizing an LLM conditioned on the dataset characteristics and user requirements, the tool selector dynamically configures the analysis pipeline, avoiding the drawbacks of static default settings and ensuring compatibility with the data's intrinsic spatiotemporal structure.

\textit{Coder}. The Coder functions as a specialized implementer, synthesizing executable scripts to instantiate abstract TI methods into runnable workflows. Leveraging a polyglot architecture, it flexibly interfaces with optimal tools across the Python and R ecosystems to resolve complex lineage topologies. The generation process adheres to rigorous constraints, managing the serialization of single-cell objects and ensuring the seamless flow of pseudotime and trajectory coordinates between heterogeneous analysis environments.

\textbf{Evaluator agent}.
The evaluator agent enforces a rigorous quality control protocol through a dual-layer validation mechanism comprising a \textit{code evaluator} for execution integrity and a \textit{biological evaluator} for scientific plausibility.

\textit{Code evaluator}. This component constitutes the foundational verification stage, specifically tasked with guaranteeing the integrity of TI workflows. It systematically inspects execution logs for algorithm-specific failures and verifies the existence of output files. If runtime exceptions are detected, the component extracts the error context and routes it as actionable debugging feedback to the executor. This enables the system to autonomously attempt code correction, explicitly skipping the biological validation phase to optimize token usage.

\textit{Biological evaluator}. The biological evaluator moves beyond syntax to assess the biological fidelity of the inferred trajectories. Ingesting the generated plots and statistical summaries, the agent utilizes LLMs to cross-reference the findings with established biological priors. It specifically targets common TI artifacts, such as biologically impossible lineage transitions or inverted developmental paths. For example, it checks alignment with the known cellular hierarchy, allowing it to flag biologically impossible transitions, such as erroneously assigning a terminally differentiated cell as the trajectory root. If such a discrepancy is detected, the biological evaluator flags the result as biologically incoherent and provides a targeted diagnostic hypothesis, guiding the planner to refine the upstream configuration.

\textbf{Knowledge-augmented fallback mechanism}. 
To enhance generalization to trajectory inference tasks outside the model's experience distribution, we design a knowledge-augmented fallback mechanism integrated into the biological evaluator agent. Specifically, when SpaCellAgent encounters a trajectory inference task that falls outside its internal experience memory, the biological evaluator initiates an automated search through PubMed to retrieve the latest biological reference, including tissue-specific markers and similar case studies. This allows the agent to evaluate its TI results in peer-reviewed scientific evidence rather than heuristic guessing.

\textbf{Dynamic tool discovery and registration mechanism}. To break through the boundaries of a predefined registry, SpaCellAgent is designed with a dynamic tool discovery and registration mechanism, enabling adaptive bioinformatics problem-solving. Specifically, if the agent determines that the existing tool registry is insufficient to execute the formulated plan, it will dynamically introduce and invoke new tools (e.g., importing a novel Python/R package for a specific spatial transcriptomics algorithm). Once the analysis incorporating the new tool successfully passes biological validation, SpaCellAgent will archive the successful analysis pipeline into the global memory and extract and register the newly utilized tool into the tool registry. Through this strategy, the tool registry autonomously expands and evolves by integrating novel tools and registering validated workflows.

\textbf{Dual-layer memory architecture}. To ensure coherent reasoning across varying temporal scales, SpaCellAgent implements a hierarchical memory system comprising local and global modules. Local memory serves as an ephemeral, intra-task context buffer. It logs the full trajectory of the current analysis step, including iterative code drafts, execution traces, and diagnostic evaluations. This comprehensive history allows the coder to engage in context-aware debugging, preventing recursive errors during self-correction. In contrast, global memory serves as a persistent inter-task knowledge base. It consolidates successfully verified code snippets and reasoning patterns from historical runs. By retrieving relevant exemplars from this global repository, the agent enables knowledge transfer across different datasets, effectively bootstrapping new tasks with proven analytical templates. This dual capability equips it with greater adaptability and scalability when confronting the high heterogeneity and complex dependencies inherent in spatial and single-cell data.
\subsection{Self-refinement and Self-evolution}

\begin{table*}[t]
  \centering
  \small
  
  \renewcommand{\arraystretch}{1.2}
  \setlength{\tabcolsep}{6pt} 
  
  \caption{Performance comparison with baseline methods on real and synthetic datasets. \textbf{Bold} indicates the best result, and \underline{underline} indicates the second-best result.}
  \label{tab:1}

  \begin{tabular}{l | cccc | cccc | cccc}
    \toprule
    
    \multirow{2}{*}{\textbf{Method}} & 
    \multicolumn{4}{c|}{\textsc{Real-Gold}} & 
    \multicolumn{4}{c|}{\textsc{Real-Silver}} & 
    \multicolumn{4}{c}{\textsc{Synthetic}} \\

    \cmidrule(lr){2-5} \cmidrule(lr){6-9} \cmidrule(lr){10-13}

    & Corr & HIM & WCor & F1 & Corr & HIM & WCor & F1 & Corr & HIM & WCor & F1 \\
    \midrule
    
    DPT & 
    0.213 & 0.515 & 0.385 & 0.262 & 
    0.259 & 0.387 & 0.413 & 0.279 & 
    0.308 & 0.483 & 0.323 & 0.345 \\
    
    RaceID/StemID & 
    0.220 & 0.334 & 0.501 & 0.206 & 
    0.288 & 0.389 & 0.501 & 0.206 & 
    0.361 & 0.381 & 0.562 & 0.128 \\
    
    Scorpius & 
    0.295 & 0.576 & 0.551 & 0.139 & 
    \underline{0.602} & \underline{0.569} & \underline{0.719} & \underline{0.574} & 
    0.455 & 0.431 & 0.525 & 0.436 \\
    
    PAGA & 
    0.256 & 0.469 & \underline{0.735} & 0.402 & 
    0.431 & 0.497 & 0.680 & 0.381 & 
    0.470 & 0.539 & 0.649 & 0.358 \\ 
    
    PAGA Tree & 
    \underline{0.319} & 0.523 & 0.478 & 0.440 & 
    0.442 & \textbf{0.614} & 0.670 & 0.519 & 
    \underline{0.501} & \textbf{0.605} & \underline{0.664} & \underline{0.439} \\
    
    Slingshot & 
    0.347 & \underline{0.651} & \textbf{0.749} & \underline{0.483} & 
    0.469 & 0.551 & 0.639 & 0.541 & 
    0.448 & 0.431 & 0.445 & 0.421 \\
    
    \midrule
    

    \rowcolor{gray!10} 
    \textbf{SpacellAgent} & 
    \textbf{0.480} & \textbf{0.710} & 0.711 & \textbf{0.608} & 
    \textbf{0.774} & 0.322 & \textbf{0.719} & \textbf{0.645} & 
    \textbf{0.564} & \underline{0.560} & \textbf{0.660} & \textbf{0.509} \\
    
    \bottomrule
  \end{tabular}
\end{table*}

\textbf{Self-refinement mechanism}.
To ensure robustness in TI analysis, SpaCellAgent incorporates an autonomous self-refinement mechanism. For each step, the system initiates a bounded iterative process in which the evaluator assesses the execution output. Successful iterations commit to the intermediate dataset state and advance the workflow. In contrast, failures trigger a refinement cycle: the coder receives an augmented context containing the full interaction history, truncated error logs, and semantic feedback from the evaluator. Guided by prompts that enforce strategy divergence to avoid repetitive errors, the agent synthesizes revised code for re-execution. This loop persists until success is achieved or the maximum attempt limit is reached. By integrating rule-based error detection with LLM-driven diagnosis, this mechanism autonomously resolved syntax errors and API discrepancies encountered during trajectory analysis, enabling the seamless modeling of cellular dynamics.

\textbf{Cross-task self-evolution}. 
Beyond within-task self-refinement, SpaCellAgent integrates a self-evolution mechanism for continual learning in TI analysis. This module treats successful TI workflows as reusable knowledge, archiving verified code snippets alongside biological contextual metadata into a global memory repository. Upon encountering new datasets, the coder queries this repository to retrieve relevant analysis templates, effectively utilizing a retrieval-augmented generation strategy to bias the model toward proven architectural patterns. Furthermore, the system accumulates error-fix pairs from historical failures, allowing it to proactively apply discovered remedies, such as parameter adjustments or alternative method selection, to recurrent issues. Over time, this cumulative knowledge base transforms SpaCellAgent into an adaptive system, progressively enhancing convergence speed, stability, and trajectory analysis consistency across diverse biological scenarios.

Together, the self-refinement and self-evolution mechanisms transform SpaCellAgent from a static pipeline into a continuously improving system. As it is applied to more datasets and use cases, its internal knowledge base becomes richer, enabling superior generalization to unseen biological contexts and robust handling of complex lineage topologies over time.

\section{Experiment}

\begin{figure*}[t]
    \centering
    \includegraphics[width=1\linewidth]{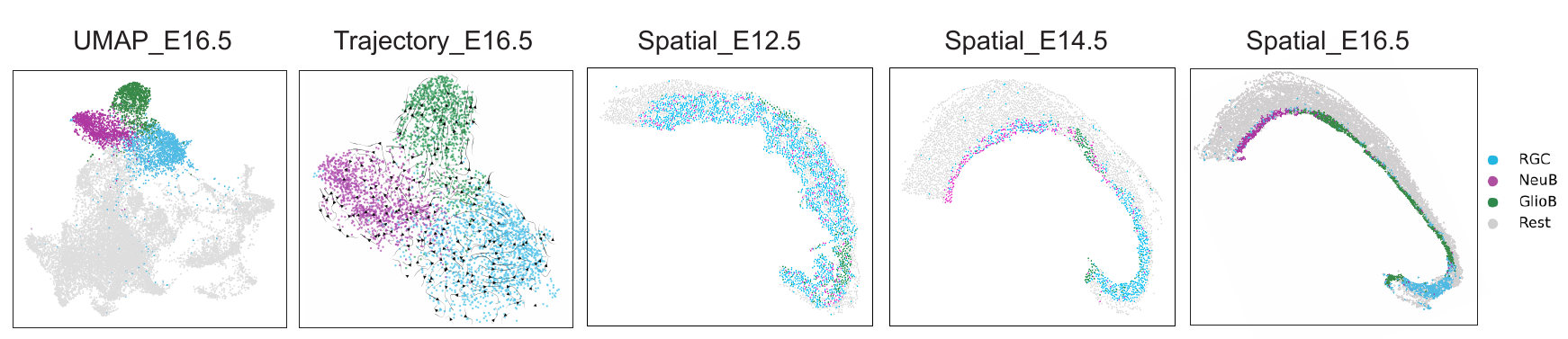}
    \caption{UMAP visualization and spatial distributions of cell types in the mouse embryonic dorsal midbrain at developmental stages E12.5, E14.5, and E16.5.}
    \Description{}
    \label{fig:3}
\end{figure*}

\subsection{Setup}
\textbf{Model configuration}. In all reported experiments, the agents were instantiated using the DeepSeek-V3 model \cite{guo2025deepseek}. We accessed the model through API with the temperature set to 0.0 and the $Top\text{-}p$ set to 1 to prioritize precise and deterministic outputs. For the self-reflection task, we employed a slightly higher temperature of 0.4 (with $Top\text{-}p$ to 1), allowing the agent to explore alternative biological hypotheses when the initial path failed. Additionally, the maximum output token limit was set to 128,000 to accommodate long-sequence biological reasoning.

\textbf{Datasets}. In this study, we employed a diverse set of six datasets comprising both synthetic and curated benchmarks (\textbf{REAL-GOLD}, \textbf{REAL-SILVER}, \textbf{SYNTHETIC}) \cite{saelens2019comparison}, and real-world spatial tissues, such as \textbf{the Mouse Dorsal Midbrain}, \textbf{Axolotl Neuron Regeneration} \cite{chen2022spatiotemporal}, and an unpublished \textbf{Mouse Spinal Cord Injury (SCI) Dataset}. The details of the datasets are in Appendix \ref{appendix:dataset}.
These datasets provide comprehensive coverage of contexts essential for robust TI evaluation: maturation, cellular states, germline specification, technical variability, and spatial organization. 

\textbf{Baselines.} We compared SpaCellAgent with five established methods representing diverse algorithmic paradigms: 
\begin{itemize} 
\item \textbf{DPT} \cite{haghverdi2016diffusion}: A \textit{diffusion-based} method that estimates pseudotime by simulating random walks on a transition matrix to resolve branching lineages.
\item \textbf{RaceID/StemID} \cite{grun2016novo}: A \textit{clustering-based} framework that combines outlier-sensitive clustering with minimum spanning trees to identify lineage connectivity. 
\item \textbf{Scorpius} \cite{cannoodt2016scorpius}: A \textit{linear inference} method that orders cells by projecting them onto a shortest-path curve derived from k-nearest neighbor distances. 
\item \textbf{PAGA} \cite{wolf2019paga}: A \textit{graph abstraction} technique that preserves global topology by constructing a coarse-grained connectivity map between cell clusters, capable of capturing complex cyclic structures. 
\item \textbf{PAGA Tree} \cite{wolf2019paga}: A variant of PAGA optimized for \textit{hierarchical lineages}, which directs the abstracted graph to generate a rooted, tree-like trajectory structure. 
\item \textbf{Slingshot} \cite{street2018slingshot}: A \textit{curve-fitting} approach that employs semi-supervised principal curves to model smooth, branching trajectories across pre-defined cluster centers. 
\end{itemize}
For all baselines, we adopt the default parameter settings as provided in their original implementations.

\subsection{Evaluation Metrics}
We employed four complementary metrics to assess SpaCellAgent from the perspectives of cellular ordering, branch assignment, biological feature relevance, and global network topology.

\textbf{Correlation}. We calculated the Spearman rank correlation coefficient $\rho$ between the geodesic distance matrices of reference and predicted trajectories:
\begin{equation}
    \text{Corr} = \rho({D}_{ref}, {D}_{pred}).
\end{equation}
This metric quantifies the preservation of global cellular relationships along the inferred milestone network.

\textbf{F1 Branches (F1)}. We used the Jaccard-based F1 score to evaluate structural correspondence:
\begin{equation}
    F1_{branch} = 2 \cdot \frac{Recovery \cdot Relevance}{Recovery + Relevance}.
\end{equation}
It assesses the model's accuracy in capturing milestone branches and ensuring the purity of cell assignments to those branches.

\textbf{Feature importance weighted correlation (wCor)}. This metric evaluates the agent's ability to identify key trajectory-driving genes. We performed Random Forest (RF) regression to extract gene importance vectors from the predicted coordinates:
\begin{equation}
    wCor = \text{corr}_w(\text{FI}_{ref}, \text{FI}_{pred}).
\end{equation}
The Pearson correlation is weighted by reference importance to emphasize biologically significant drivers.

\textbf{Hamming-Ipsen-Mikhailov score (HIM)}. To compare the inferred milestone network $\mathcal{G}_{pred}$ with the ground truth $\mathcal{G}_{ref}$, we employed the HIM distance:
\begin{equation}
    HIM = 1- \frac{1}{\sqrt{2}} \sqrt{d_H^2 + d_{IM}^2},
\end{equation}
where $d_H$ represents normalized Hamming distance and $d_{IM}$ represents Ipsen-Mikhailov distances. 

For all metrics defined above, higher values consistently indicate superior performance.

\subsection{Results}
\begin{figure*}[t]
    \centering
    \includegraphics[width=1\linewidth]{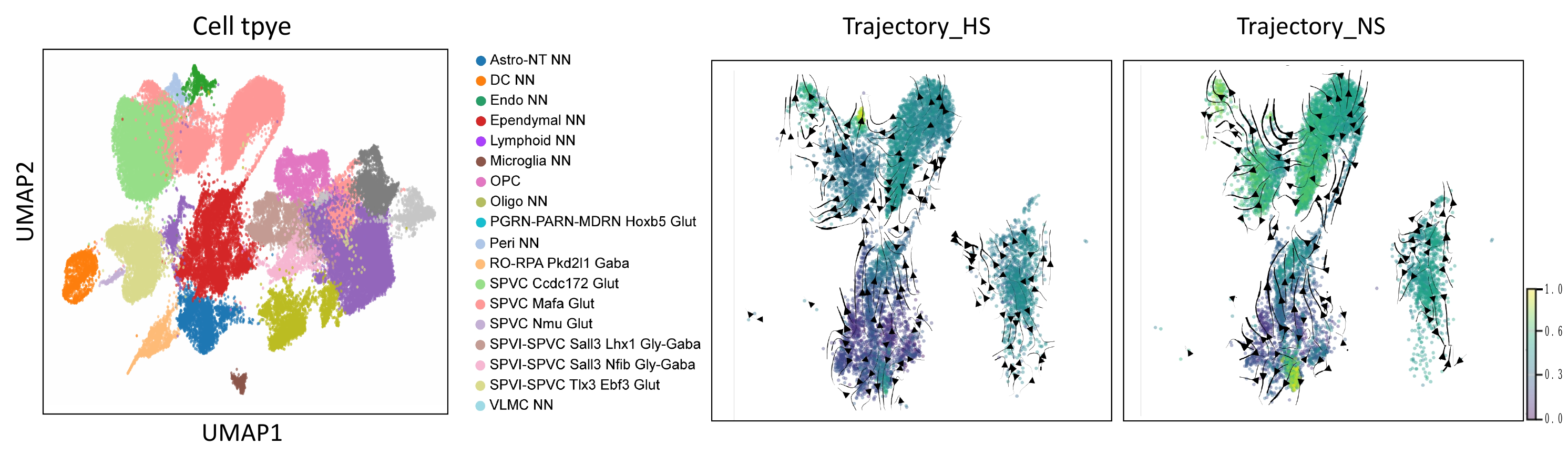}
    \caption{UMAP projection of cell type clusters identified by SpaCellAgent and comparison of inferred developmental trajectories between Homeostatic (HS) and Nerve Injury (NS) conditions.}
    \Description{}
    \label{fig:4}
\end{figure*}
To evaluate the trajectory analysis capabilities of SpaCellAgent, we first evaluated both baselines and SpaCellAgent across three diverse benchmarks: \textbf{REAL-GOLD}, \textbf{REAL-SILVER}, and \textbf{SYNTHETIC}. The general results shown in Table \ref{tab:1} present a comprehensive performance comparison. Then, we conducted a comparison with LLM-based baselines. The details are provided in Appendix \ref{appendix:addexper}.

As shown in Table \ref{tab:1}, SpaCellAgent achieves SOTA performance, ranking first in the majority of metrics across all datasets. Notably, on the REAL-GOLD dataset, SpaCellAgent outperforms the second-best method, Slingshot, with an average Correlation of 0.480 and an F1 score of 0.608. This substantial improvement is attributed to the self-refinement and self-evolution mechanisms within SpaCellAgent, which autonomously mitigate errors arising from manual parameter tuning. In contrast to baseline methods constrained by static algorithmic assumptions, SpaCellAgent employs its LLM-driven planner and tool selector to perform dynamic data profiling. This mechanism empowers the agent to autonomously orchestrate the optimal configuration of analytical tools tailored to each specific dataset, thereby circumventing the drawbacks of human bias and manual optimization errors. 

\textbf{Spatial trajectory analysis and biological insights.} To further investigate the capability of SpaCellAgent in analyzing high-dimensional data with spatial coordination, we applied SpaCellAgent to the Mouse Embryonic Dorsal Midbrain dataset. This dataset contains 26,738 spatially resolved cells from the mouse embryonic dorsal midbrain, with expression profiles for 24,045 genes and temporal information from stages E12.5, E14.5, and E16.5. Unlike dissociated single-cell data, this task requires the agent to reconcile gene expression similarity with physical proximity.
\begin{figure}
    \centering
    \includegraphics[width=1\linewidth]{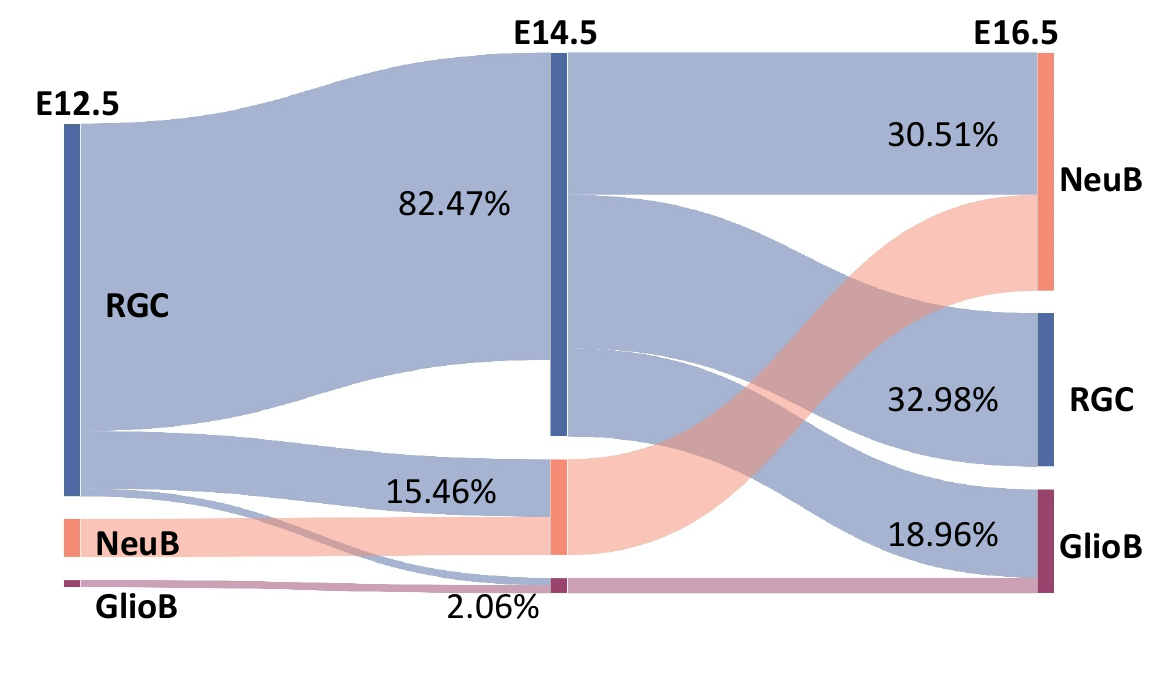}
    \caption{Sankey diagram illustrating the temporal lineage transitions and cell fate proportions across the three embryonic stages.}
    \Description{}
    \label{fig:5}
\end{figure}

\begin{figure}[t]
    \centering
    \includegraphics[width=0.73\linewidth]{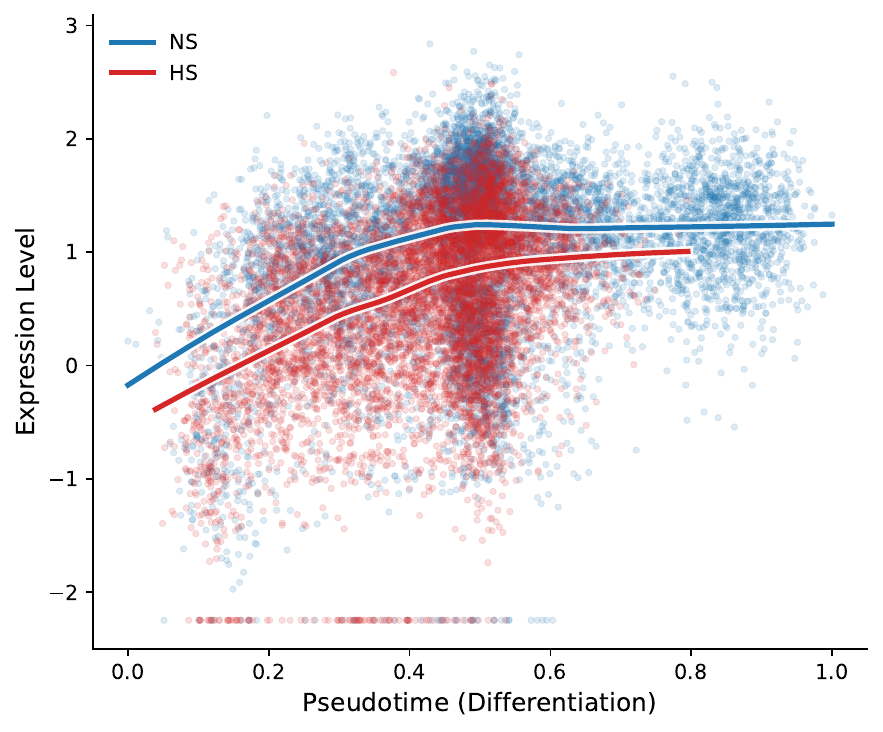}
    \caption{Pseudotime expression kinetics of the marker gene Plp1, showing differential differentiation trends between HS and NS groups. Solid lines represent the fitted expression trends along the differentiation pseudotime.}
    \Description{}
    \label{fig:6}
\end{figure}

As shown in Figure \ref{fig:4}, SpaCellAgent autonomously orchestrated the data preprocessing and visualized the clustering landscape through UMAP. Critically, SpaCellAgent proactively inferred the latent developmental pathways and automatically generated trajectory plots based on its reasoning. These visualizations reveal that SpaCellAgent accurately reconstructed the bifurcation trajectory from Radial Glia-like (RGL) progenitors into distinct neuronal (NeuB) and glial (GlioB) lineages. This trajectory aligns with the temporal progression from E12.5 to E16.5, as illustrated in Figure \ref{fig:5}, and is driven by a distinct shift in marker expression, transitioning from \textit{Sox2}-positive progenitors to \textit{Nurr1}- and \textit{Th}-positive neurons. These results are consistent with prior studies \cite{chen2022spatiotemporal,kee2017single}. Subsequently, we conducted functional enrichment analysis, validating the biological plausibility of the identified lineage, highlighting key pathways such as axonogenesis and dopamine metabolic processes, which are intrinsic to midbrain development. Next, we applied SpaCellAgent to the Axolotl Neuron Regeneration dataset, another spatially resolved dataset capturing complex tissue reconstruction. 
%

\subsection{Case Study}

To evaluate the generalization capability and real-world utility of SpaCellAgent, we deployed the framework on a private scRNA-seq dataset of Mouse SCI provided by our collaborating hospital. This dataset captures the impact of high-salt (HS) versus normal-salt (NS) intake across three specific time points: Day 0, Day 7, and Day 28. Specifically, time point day 0 denotes the Sham-operated (SAM) group, serving as the uninjured baseline control where the surgical procedure was performed without inducing spinal cord damage.

SpaCellAgent first devised a TI analysis plan based on the structural characteristics of the data and user requirements, and designated the optimal analytical tool for each step of the workflow. Subsequently, SpaCellAgent executed the workflow according to the specified plan. Following the cell annotation step, it analyzed potential cell developmental trajectories within the SCI context based on the annotation results. Subsequently, SpaCellAgent executed the TI step, modeling the lineage progression from precursors to mature cells by constructing a pseudo-temporal ordering, as shown in Figure \ref{fig:4}. SpaCellAgent further conducted temporal expression profiling, revealing that HS-OPCs persistently upregulated the inhibitor Sox17 even at late stages while failing to sustain Plp1. 

To quantify this developmental perturbation, SpaCellAgent further employed pseudotime analysis to dissect the underlying lineage dynamics. As shown in Figure \ref{fig:6}, solid lines representing fitted expression trends along the differentiation pseudotime clearly reveal a premature termination of the HS trajectory and a significant expression gap in \textit{Plp1} compared to the NS lineage. The results reveal that, by integrating velocity-derived streamlines with fate probability calculations, the agent autonomously identified a marked differentiation blockade in the HS group. Unlike the continuous NS trajectory, the HS lineage exhibited fragmented streamlines and early-stage cell arrest, indicating impaired endogenous remyelination. These findings are consistent with prior studies highlighting the detrimental impact of high-salt diets on post-injury neural recovery \cite{kleinewietfeld2013sodium,wu2013induction}. These results substantiate SpaCellAgent’s capacity for autonomous hypothesis formulation in unseen biological contexts, proving its practical utility as a powerful assistant for accelerating novel scientific discoveries. Then, we conducted an enrichment analysis to quantitatively validate the biological fidelity. The details are provided in Appendix \ref{appendix:addexper}.

\section{Ablation Studies}

\begin{figure}
    \centering
    \includegraphics[width=0.8\linewidth]{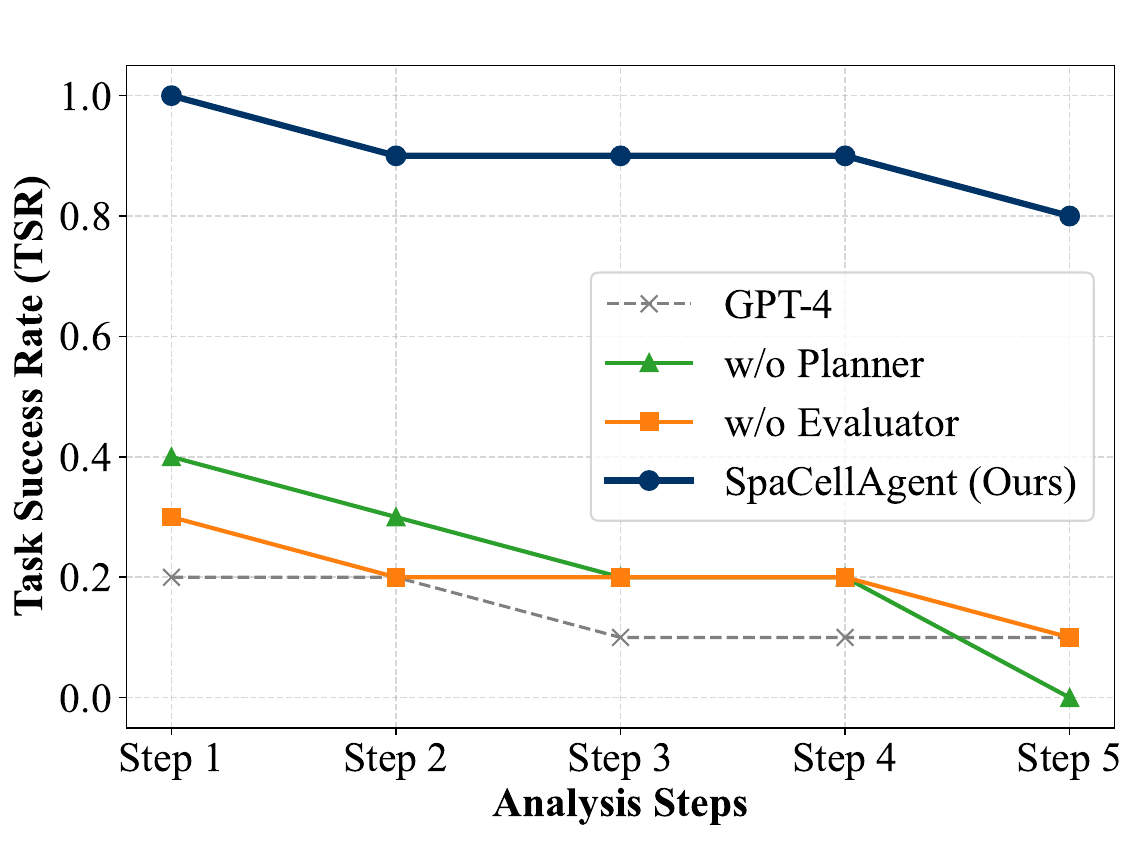}
    \caption{The results of operational effectiveness analysis.}
    \Description{}
    \label{fig:7}
\end{figure}

To verify the effectiveness of SpaCellAgent’s components, we conducted ablation studies with the following variants: (1) SpaCellAgent w/o planner; (2) SpaCellAgent w/o evaluator; (3) SpaCellAgent w/o self-evolution (w/o Evol.); (4) GPT-4 (w/o agent); (5) w/o experience memory. 

\textbf{Effectiveness analysis}. We first conducted ablation experiments on the REAL-GOLD benchmark, alongside the Mouse Embryonic Dorsal Midbrain and Axolotl Neuron Regeneration Datasets. For this analysis, we defined a standardized TI workflow comprising six tasks and measured the Task Success Rate (TSR) to compare different variants. The results given in Figure \ref{fig:7} show the necessity of the full multi-agent architecture. Specifically, GPT-4 exhibits the poorest performance, primarily due to its inability to manage the intricate context of long-horizon TI workflows, which often results in code generation failures. Regarding the architectural variants, the w/o planner variant suffers from a lack of strategic direction, leading to failures in decomposing complex biological goals into actionable sub-tasks. Similarly, the w/o evaluator variant operates without the critical self-correction mechanism, rendering it incapable of resolving runtime errors or refining suboptimal outputs, which significantly undermines overall robustness compared to the SpaCellAgent framework.

\textbf{Efficiency analysis}. To evaluate efficiency, we recruited five domain experts actively engaged in bioinformatics research (spanning Master's and Ph.D. levels) to establish an averaged human baseline. The details of experts team are provided in Appendix \ref{abdetail}. Then, we benchmarked SpaCellAgent against the human baseline and the variant w/o self-evolution. As shown in Table \ref{tab:efficiency_comparison}, SpaCellAgent recorded the lowest total execution time of 38.0 minutes, outperforming the human average by 41.2\% and the non-evolutionary baseline by 32.7\%. Gains were most significant in complex downstream tasks, with SpaCellAgent surpassing human experts by 53.6\% in Step 4 and 63.9\% in Step 5. While humans slightly lead in initial preprocessing, the self-evolution mechanism minimizes redundant debugging in later stages, ensuring superior scalability for end-to-end analysis.
Collectively, these results demonstrate that SpaCellAgent significantly alleviates the manual burden of complex trajectory inference, offering a scalable and time-efficient paradigm for high-throughput biological hypothesis formulation.

\textbf{Cross-Model sensitivity analysis}.
To demonstrate our framework's robustness, we conducted additional sensitivity experiments using GPT-5.2 and claude-sonnet-4-6 as alternative foundational models. We experimented with five subsets from the real gold benchmark and calculated Spearman's rank correlation. Meanwhile, we conducted an effectiveness analysis on complex real-world datasets for 10 runs and calculated the success rate (SR). The results of these cross-model comparisons are shown in Table \ref{tab: Sensitivity}. The results show consistent gains in both efficiency and accuracy, confirming that the performance is driven by our agentic framework design.

\begin{table}
\centering
\caption{Efficiency comparison, time cost in minutes.}
\label{tab:efficiency_comparison}
\resizebox{\columnwidth}{!}{%
\begin{tabular}{l c c c}
\toprule
\textbf{Task} & \textbf{Human Expert} & \textbf{w/o Evol.} & \textbf{SpaCellAgent} \\
\midrule
Step1 & \textbf{10.0 ± 1.6} & 15.4 ± 1.2 & 12.4 ± 0.8 \\
Step2 & 8.4 ± 1.2 & 6.8 ± 2.0 & \textbf{6.5 ± 1.4} \\
Step3 & 11.2 ± 2.0& 8.8 ± 0.8 & \textbf{5.6 ± 0.6} \\
Step4 & 8.4 ± 1.2 & 11.0 ± 1.2 & \textbf{3.9 ± 0.7} \\
Step5 & 26.6 ± 3.8 & 14.5 ± 1.5 & \textbf{9.6 ± 0.9} \\
\midrule
\textbf{Total Time} & 64.6 & 56.5 & \textbf{38.0} \\
\bottomrule
\end{tabular}%
}
\end{table}

\begin{table}
\centering
\caption{Sensitivity comparison with different LLMs.}
\label{tab: Sensitivity}
\begin{tabular}{l c c c c c c}
\toprule
\textbf{Dataset} & \textbf{Deepseek-V3} & \textbf{GPT-5.2} & \textbf{claude-sonnet-4-6} \\
\midrule
S1 & 0.526 & 0.538 & 0.518 \\
S2 & 0.615 & 0.613 & 0.624 \\
S3 & 0.571 & 0.570 & 0.582 \\
S4 & 0.483 & 0.494 & 0.481 \\
S5 & 0.755 & 0.768 & 0.768 \\
\midrule
SR & 0.8 & 0.9 & 0.9 \\
\bottomrule
\end{tabular}%
\end{table}

\textbf{Tool selection consistency analysis}. To evaluate the tool selection consistency and planning accuracy of SpaCellAgent, we conducted systematic experiments of potential issues such as tool selection errors and agent planning failures. Specifically, we repeated the same analysis task 5 times independently on each of the three real-world datasets: the Mouse Dorsal Midbrain (Midbrain), Axolotl Neuron Regeneration (Axolotl), and SCI (with the same input and the temperature set to 0). We measured the tool selection consistency of each dataset. Meanwhile, we invited a human expert team to evaluate the planning accuracy and feasibility (use a quality score rated on a scale from 0 to 10). We implemented a standardized TI workflow as defined in our effectiveness analysis, measuring the consistency rate across all repeated trials. Subsequently, we invited the human expert team to independently evaluate the accuracy and feasibility of the generated analysis plans and each step of the code. The results are shown in Table \ref{tab: tool}. The result shows that our framework achieves consistently high tool selection reliability and planning accuracy across all three datasets, confirming that the standardized workflow effectively mitigates tool selection errors and planning failures. 

\begin{table}
\centering
\caption{Tool selection consistency of the three real-world datasets.}
\label{tab: tool}
\begin{tabular}{l c c c c c c}
\toprule
\textbf{Dataset} & \textbf{Midbrain} & \textbf{Axolotl} & \textbf{SCI} \\
\midrule
Step1 & 1.0 & 1.0 & 1.0 \\
Step2 & 1.0 & 1.0 & 1.0 \\
Step3 & 1.0& 1.0 & 1.0 \\
Step4 & 1.0 & 1.0 & 1.0 \\
Step5 & 0.9 & 0.8 & 1.0 \\
\midrule
\textbf{Quality score} & 8.0 ± 0.7 & 7.8 ± 0.4& 7.2 ± 0.8 \\
\bottomrule
\end{tabular}%
\end{table}

To quantitatively verify the effectiveness of the Knowledge-augmented fallback mechanism, we conduct an additional ablation study with the w/o experience memory variants on the real-gold dataset. The details are provided in Appendix \ref{appendix_addab}.
\section{Conclusion}

We introduce SpaCellAgent, an autonomous self-evolving LLM-based multi-agent framework for trajectory analysis. By integrating a collaborative multi-agent architecture with dynamic tool orchestration, SpaCellAgent plans workflows, selects methods, executes analyses, and generates interpretable reports, closing the gap between raw data and biological insight. The results of the experiment confirm that the SpaCellAgent matches the performance of the experts while reducing the analysis time by more than 40\% and lowering the accessibility barrier for the domain researchers. Ablation studies validate the critical roles of agent-based decomposition, iterative self-refinement, and knowledge reuse across tasks in achieving robustness and stability. Future work will extend the SpaCellAgent to spatial multi-omics integration and tighter coupling with regulatory network analysis, positioning SpaCellAgent as a foundational step toward scalable, intelligent assistants in computational biology.

\section*{Acknowledgments}
This work was partially supported by the National Natural Science Foundation of China (Grant No. 62525213, 62372459, 62325604), the Beijing Natural Science Foundation (Grant No. L253021), and the Innovative Research Group Project of Hunan Province (Grant No. 2025JJ10008).
\section*{Limitations and Ethical Considerations}
All authors adhere to all ACM Publications Policies, including ACM's Publications Policy on Research Involving Human Participants and Subjects. The regulatory mechanisms inferred by SpaCellAgent, while computationally robust and consistent with established literature, remain hypotheses that warrant future validation through in vivo experimental assays. All datasets utilized in this study were acquired and processed in strict adherence to ethical guidelines and privacy regulations.

\section*{GenAI Disclosure}
In this work, we mainly employ GenAI to identify and correct syntax errors and polish writing. 

\bibliographystyle{ACM-Reference-Format}
\balance
\bibliography{reference}

\newpage
\appendix

\section{Methods Details}

\subsection{TI analysis}
\label{appendix:ti_formalization}

We formalize the Trajectory Inference (TI) task as a manifold learning problem. Given a single-cell gene expression matrix $\mathbf{X} \in \mathbb{R}^{N \times M}$ comprising $N$ cells and $M$ genes, the objective is to learn a low-dimensional manifold $\mathcal{M}$ embedded in a latent space $\mathcal{Z} \in \mathbb{R}^{N \times d}$ (where $d \ll M$), and a pseudotime function $\tau: \mathcal{Z} \rightarrow \mathbb{R}_{\ge 0}$ that maps each cell to its developmental progression. We contrast two dominant formalisms employed in our evaluation: cluster-based abstraction and principal graph learning.

\subsubsection{Cluster-based Formalism}
This class of methods approximates the continuous manifold $\mathcal{M}$ using a coarse-grained graph derived from discrete cell partitions.

\textbf{Manifold Partitioning:} The latent space $\mathcal{Z}$ is partitioned into $k$ disjoint clusters $\mathcal{C} = \{C_1, C_2, \dots, C_k\}$ using algorithms such as Louvain or $k$-means. Each cluster is represented by a centroid $\mu_i$.
    
\textbf{Graph extraction:} A connectivity graph $\mathcal{G}_{abs} = (\mathcal{V}, \mathcal{E})$ is constructed where nodes $\mathcal{V}$ correspond to cluster centroids. An edge $e_{ij} \in \mathcal{E}$ is established if the connectivity metric $S(C_i, C_j)$ (e.g., statistical interconnectivity in PAGA) exceeds a threshold $\delta$:
    \begin{equation}
        \mathcal{E} = \{ (i, j) \mid S(C_i, C_j) > \delta \}.
    \end{equation}
    
\textbf{Curve Fitting and Pseudotime:} A set of smooth principal curves $\Gamma = \{\gamma_1, \dots, \gamma_b\}$ is fitted through the centroids along the topology of $\mathcal{G}_{abs}$. The pseudotime $\tau_i$ for a cell $\mathbf{z}_i$ is computed by projecting the cell onto the nearest curve $\gamma$:
    \begin{equation}
        \tau_i = \int_{\text{root}}^{\text{proj}(\mathbf{z}_i, \gamma)} || \frac{\partial \gamma(t)}{\partial t} || dt.
    \end{equation}

\subsubsection{Principal Graph Learning}
Instead of relying on discrete clusters, this formalism employs \textit{Reverse Graph Embedding} (RGE) to learn a continuous tree structure $\mathcal{T}$ directly within the latent space.

\textbf{Graph Initialization:} A simpler graph structure (e.g., a tree) is initialized in the latent space $\mathcal{Z}$.
    
\textbf{Optimization Objective:} The algorithm iteratively updates the graph nodes $\mathcal{V}_{\mathcal{T}}$ and edges to minimize the distance between the data points and the graph structure, typically minimizing a regularized quantization error:
    \begin{equation}
        \min_{\mathcal{T}, \mathbf{Y}} \sum_{i=1}^{N} || \mathbf{z}_i - f(\mathbf{z}_i) ||^2 + \lambda \Omega(\mathcal{T}).
    \end{equation}
    where $f(\mathbf{z}_i)$ projects cell $i$ to the nearest point on the graph $\mathcal{T}$, and $\Omega(\mathcal{T})$ is a regularization term enforcing graph smoothness or tree structure.
    
\textbf{Geodesic Pseudotime:} Upon convergence, the pseudotime is defined as the geodesic distance $d_g(\cdot, \cdot)$ along the learned principal graph from a designated root node $r$:
    \begin{equation}
        \tau_i = d_g(\text{proj}(\mathbf{z}_i, \mathcal{T}), r).
    \end{equation}

\subsection{System Prompts and Instructions}
\label{appendix:Prompts}
The cognitive architecture of SpaCellAgent is governed by role-specific system instructions that delineate the operational boundaries and reasoning protocols for each agent. Below, we provide the \textbf{verbatim system prompts} used to instantiate the Planner, Executor, and Evaluator agents. These prompts are engineered to enforce strict adherence to biological logic, code syntax correctness, and iterative self-refinement.

\label{appendix:Prompts}
The cognitive architecture of SpaCellAgent is governed by role-specific system instructions that delineate the operational boundaries and reasoning protocols for each agent. Below, we provide the \textbf{verbatim system prompts} used to instantiate the Planner, Executor, and Evaluator agents. These prompts are engineered to enforce strict adherence to biological logic, code syntax correctness, and iterative self-refinement.

\subsubsection{Planner Agent Prompt}
The Planner is instructed to decompose the user's biological query into a logical sequence of analytical milestones.
\begin{figure}
 
\begin{promptbox}[Planner Agent Output Format]
{
  "User Task Description": "{user_task}",
  "User Data Description": "{data_representation}",
  "Output Format Requirements": {
    "steps": [
      {
        "id": 1,
        "description": "Step 1 description"
      },
      {
        "id": 2,
        "description": "Step 2 description"
      }
    ]
  }
}
\end{promptbox}
 \Description{}  
\end{figure}

\subsubsection{Executor Agent Prompt}
The Executor is tasked with translating the high-level plan into executable Python scripts, utilizing the predefined tool registry.
\begin{figure*}[t!]
    \centering

    \begin{promptbox}[Executor Agent System Prompt (Part I: Context)]
{
  "prompt": [
    "You are a professional bioinformatics code writing expert.",
    "Please write Python code to complete the current task step based on the following information.",
    
    "User Requirements:",
    "{user_requirements}",
    "Data Description:",
    "{data_description}",
    "Historical Code:",
    "{historical_code}",
    "Current Task Step:",
    "{step_description}"
  ]
}
    \end{promptbox}
    
    \Description{System prompt part 1 showing context inputs.}
    \label{fig:prompt_part1}
\end{figure*}

\begin{figure*}[t!]
    \centering

    \begin{promptbox}[Executor Agent System Prompt (Part II: Requirements)]
{
  "prompt_continued": [
    "Selected Tools and Their Documentation:",
    "{tools_docs}",
    "**Important: Data Loading Strategy**",
    "{data_loading_strategy}",
    
    "Requirements:",
    "- Use the selected tools to complete the task.",
    "- The code should include necessary imports and data loading.",
    "- **Data Loading**: {data_loading_instruction}",
    

    "- **Data Saving**: At the end of the code, you must add code to save intermediate",
    "  results: `adata.write_h5ad('{intermediate_file_path}')`",

    
    "- The code should use the data file path: {data_file_path}",
    "- **Must use correct Python syntax**",

    "- If the current step involves \"trajectory inference / pseudotime...\",", 
    "  please follow these additional output specifications:",
    "{ti_output_requirements}",

    
    "- Do not add any explanations or comments, only provide code.",
    "- The output format is a code block, wrapped with ```python and ```."
  ]
}
    \end{promptbox}
    \Description{System prompt part 2 showing execution requirements.}
    \label{fig:prompt_part2}
\end{figure*}
\subsubsection{Evaluator Agent Prompt}
The Evaluator serves as the quality assurance module, analyzing execution logs and visual outputs to synthesize refinement strategies.
\begin{figure}

\begin{promptbox}[Evaluator Agent System Prompt]
{
  "prompt": [
    "You are an experienced bioinformatics expert.",
    "Please evaluate whether the code execution was successful...",
    "",
    "**Important Evaluation Rules**:",
    "1. If the execution results contain any of the following keywords, the execution failed:",
    "   - \"Execution error\", \"Execution failed\"",
    "   - \"Error\", \"Exception\", \"Traceback\"",
    "   ...",
    "",
    "User Requirements:",
    "{user_requirements}",
    "",
    "Please output strictly in the following format..."
  ]
}
\end{promptbox}
\Description{}  
\end{figure}
\begin{figure*}[t]
\begin{promptbox}[Bio-Evaluator Agent System Prompt]
{
  "prompt": [
    "You are a distinguished Senior Computational Biologist.",
    "Your task is to evaluate the **scientific validity** and **biological plausibility** of the analysis results generated by the previous step.",
    "Unlike the Code Evaluator, you assume the code executed without errors; your focus is on the **quality and meaning** of the output.",
    "**Evaluation Criteria**:",
    "1. **Data Integrity & Statistics**:",
    "   - Check if the output contains valid biological signals (e.g., variable genes found, clusters identified).",
    "   - Fail if results are empty, trivial (e.g., 0 cells remaining), or statistically insignificant (e.g., all p-values > 0.05).",
    "2. **Biological Plausibility**:",
    "   - **Clustering**: Fail if the number of clusters is extreme (e.g., 1 cluster or >100 clusters for a small dataset).",
    "   - **Annotation**: Fail if marker genes are non-specific (e.g., mitochondrial genes only).",
    "3. **Trajectory Inference (TI) Specifics** (if applicable):",
    "   - **Topology**: Fail if the inferred trajectory is highly fragmented (disconnected manifolds) or purely linear when branching is expected.",
    "   - **Pseudotime**: Check if the root cell selection aligns with user intent (e.g., time 0 should be at the stem/progenitor stage).",
    "User Requirements: {user_requirements}",
    "Data Description: {data_description}",
    "Current Task Step: {step_description}",
    "Analysis Results (Logs & Summaries): {execution_result}",
    "Please output strictly in the following format:",
    "Evaluation Result: [Pass / Fail / Needs Refinement]",
    "Biological Critique: [Briefly analyze the biological meaning. E.g., \"Trajectory successfully captures the bifurcating lineage.\"]",
    "Refinement Strategy: [If Fail: Provide specific parameter adjustments (e.g., \"Increase resolution to 0.8\"). If Pass: \"None\"]"
  ]
}
\end{promptbox}
\Description{System prompt for the Bio-Evaluator Agent}  
\end{figure*}

\subsection{Tool Definition}
To facilitate the autonomous orchestration of complex bioinformatics workflows, we provide the Executor agent with a rigorous \textbf{Tool Registry Schema}. As presented in Table \ref{tab:bioinfo_tools}, each computational tool (e.g., functions from \textit{Scanpy}, \textit{Squidpy}, or \textit{Monocle3}) is encapsulated with a semantic description, strict parameter constraints, and input/output specifications. This structured definition enables the Large Language Model (LLM) to accurately map high-level analytical intent to specific function calls without hallucinating non-existent parameters.

\begin{table*}[htbp]
    \centering
    \caption{Summary of mainstream tools and workflows for single-cell and spatial transcriptomics analysis.}
    \label{tab:bioinfo_tools}
    \renewcommand{\arraystretch}{1.3} 
    
\begin{tabular}{
    >{\raggedright\arraybackslash}p{0.20\linewidth} 
    >{\raggedright\arraybackslash}p{0.45\linewidth} 
    >{\raggedright\arraybackslash}p{0.20\linewidth}  
    c                                                
}
        \toprule
        \textbf{Analysis Step} & \textbf{Description} & \textbf{Mainstream Tools} & \textbf{Lang.} \\ 
        \midrule
        
        \textbf{Preprocessing \& QC} & 
        Quality control, doublet removal, mitochondrial gene filtering, and data normalization. & 
        Scanpy, Seurat, \newline DoubletFinder & 
        Py/R \\ 
        
        \textbf{Dimensionality Reduction} & 
        Mapping high-dimensional gene expression data into low-dimensional space (PCA, UMAP, t-SNE). & 
        Scanpy, Seurat & 
        Py/R \\ 
        
        \textbf{Clustering \& Annotation} & 
        Identifying cell subgroups based on profiles and annotating cell types using marker genes. & 
        Leiden, Louvain, \newline SingleR & 
        Py/R \\ 
        
        \textbf{Trajectory Inference} & 
        Constructing pseudo-temporal ordering to reconstruct cell differentiation lineages. & 
        Monocle3, Slingshot, \newline PAGA, Palantir & 
        R/Py \\ 
        
        \textbf{RNA Velocity} & 
        Predicting future cell states using the ratio of spliced to unspliced transcripts. & 
        scVelo, velocyto & 
        Py \\ 
        
        \textbf{Fate Probability} & 
        Integrating velocity and trajectory data to compute differentiation probabilities. & 
        CellRank & 
        Py \\ 
        
        \textbf{Differential Expression} & 
        Identifying statistically significant differentially expressed genes (DEGs) between clusters. & 
        Wilcoxon Rank Sum, \newline DESeq2, MAST & 
        R/Py \\ 
        
        \textbf{Functional Enrichment} & 
        Analyzing biological pathways (GO, KEGG) enriched in specific gene sets. & 
        clusterProfiler, \newline gseapy, GSEA & 
        R/Py \\ 
        
        \textbf{Cell-Cell Communication} & 
        Inferring intercellular interaction networks based on ligand-receptor databases. & 
        CellChat, CellPhoneDB, \newline NicheNet & 
        R/Py \\ 
        
        \textbf{Spatial Analysis} & 
        Identifying spatial domains and spatially variable genes (SVGs) using coordinates. & 
        Squidpy, Giotto, \newline SpaceRanger & 
        Py/R \\ 
        \bottomrule
    \end{tabular}
\end{table*}

\section{Experimental Details}
\subsection{Dataset Details}
\label{appendix:dataset}
To comprehensively evaluate the generalization capability and robustness of the \textit{SpaCellAgent} framework, we curated a diverse benchmark suite comprising three published real-world datasets, one unpublished real-world dataset, and one simulated dataset. These datasets encompass a wide spectrum of biological contexts, ranging from distinct organisms to varying tissue types. We fix the seed as 42 to ensure that the same spatial transcriptomics dataset yields consistent analytical trajectories across multiple runs.

Crucially, the selected datasets exhibit significant heterogeneity in terms of data sparsity, sequencing depth, and underlying topological complexity (including linear, bifurcation, and multifurcation trajectories). This diversity allows us to rigorously assess the agent's ability to autonomously adapt its analysis strategy to different data distributions and sequencing technologies (e.g., 10x Genomics, Smart-seq2, and spatial transcriptomics protocols).
\begin{table*}[t]
  \centering
  \caption{Statistics of the two spatial datasets and the unpublished dataset used in this study. The symbol ``$\times$'' represents the dimensions of the expression matrix (Cells $\times$ Genes).}
  \label{tab:dataset_stats}
  \renewcommand{\arraystretch}{1.1} 
  
  \setlength{\tabcolsep}{0pt}       
  
  \begin{tabular*}{\textwidth}{l @{\extracolsep{\fill}} l c c c c}
    \toprule
    \textbf{Dataset} & \textbf{Organism} & \textbf{Shape (Cell $\times$ Gene)} & \textbf{Technology} & \textbf{Type} & \textbf{Source} \\
    \midrule
    
    Mouse Midbrain (E12.5) & Mouse & $3,671 \times 24,045$  & Stereo-seq & Spatial & Chen \textit{et al.} \\
    Mouse Midbrain (E14.5) & Mouse & $3,648 \times 24,045$  & Stereo-seq & Spatial & Chen \textit{et al.} \\
    Mouse Midbrain (E16.5) & Mouse & $19,419 \times 24,045$ & Stereo-seq & Spatial & Chen \textit{et al.} \\
    \textit{Mouse Midbrain (Total)} & \textit{Mouse} & \textit{26,738 $\times$ 24,045} & \textit{Stereo-seq} & \textit{Spatial} & Chen \textit{et al.} \\
    
    \midrule
    
    Axolotl Neuron (D2)  & Axolotl & $518 \times 10,000$  & Stereo-seq & Spatial & Chen \textit{et al.} \\
    Axolotl Neuron (D5)  & Axolotl & $726 \times 10,000$  & Stereo-seq & Spatial & Chen \textit{et al.} \\
    Axolotl Neuron (D10) & Axolotl & $911 \times 10,000$  & Stereo-seq & Spatial & Chen \textit{et al.} \\
    Axolotl Neuron (D20) & Axolotl & $1,437 \times 10,000$ & Stereo-seq & Spatial & Chen \textit{et al.} \\
    Axolotl Neuron (D30) & Axolotl & $1,345 \times 10,000$ & Stereo-seq & Spatial & Chen \textit{et al.} \\
    \textit{Axolotl Neuron (Total)} & \textit{Axolotl} & \textit{4,937 $\times$ 10,000} & \textit{Stereo-seq} & \textit{Spatial} & Chen \textit{et al.} \\
    
    \midrule
    
    Mouse SCI (HSSAM) & Mouse & $9,094 \times 55,401$  & 10x Genomics & scRNA-seq & \textbf{This work} \\
    Mouse SCI (HSD7)  & Mouse & $13,526 \times 55,401$ & 10x Genomics & scRNA-seq & \textbf{This work} \\
    Mouse SCI (HSD28) & Mouse & $11,268 \times 55,401$ & 10x Genomics & scRNA-seq & \textbf{This work} \\
    Mouse SCI (NSSAM) & Mouse & $9,926 \times 55,401$  & 10x Genomics & scRNA-seq & \textbf{This work} \\
    Mouse SCI (NSD7)  & Mouse & $8,145 \times 55,401$  & 10x Genomics & scRNA-seq & \textbf{This work} \\
    Mouse SCI (NSD28) & Mouse & $11,074 \times 55,401$ & 10x Genomics & scRNA-seq & \textbf{This work} \\
    \textit{Mouse SCI (Total)} & \textit{Mouse} & \textit{63,033 $\times$ 55,402} & \textit{10x Genomics} & \textit{scRNA-seq} & \textbf{This work} \\
    
    \bottomrule
  \end{tabular*}
\end{table*}

\begin{table*}[t]
  \centering

  \renewcommand{\arraystretch}{1.1} 
  \setlength{\tabcolsep}{15pt}
  
  \caption{Overview of the single-cell datasets selected from the REAL-GOLD and REAL-SILVER benchmarks. Columns indicate the organism, dataset shape, and experimental technology.}
  \label{tab:sc_datasets_part1}
  \begin{tabular}{l  c l l r}
    \toprule
    \textbf{Organism} & \textbf{Shape (Cell $\times$ Gene)} & \textbf{Technology} & \textbf{Type} & \textbf{Source} \\
    \midrule
    
    Mouse & $291 \times 3,582$   & Fluidigm C1 & scRNA-seq & GSE52529 \\
    Mouse & $65 \times 1,095$    & Fluidigm C1 & scRNA-seq & GSE52583 \\
    Mouse & $541 \times 3,630$   & Smart-seq   & scRNA-seq & GSE48968 \\
    Mouse & $408 \times 3,387$   & Smart-seq   & scRNA-seq & \\
    Mouse & $435 \times 3,604$   & Smart-seq   & scRNA-seq & \\
    Mouse & $264 \times 5,310$   & Fluidigm C1 & scRNA-seq & E-MTAB-2805 \\
    
    \midrule
    
    Human & $272 \times 5,708$   & Tang \textit{et al.} & scRNA-seq & GSE63818 \\
    Human & $166 \times 3,457$   & Tang \textit{et al.} & scRNA-seq & \\
    Human & $101 \times 3,502$   & Tang \textit{et al.} & scRNA-seq & \\
    Human & $222 \times 3,766$   & Fluidigm C1 & scRNA-seq & GSE64016 \\
    Mouse & $238 \times 1,845$   & Fluidigm C1 & scRNA-seq & GSE60781 \\
    Human & $1,299 \times 4,135$ & Fluidigm C1 & scRNA-seq & E-MTAB-3929 \\
    
    \midrule
    
    Mouse & $873 \times 2,863$   & Smart-seq   & scRNA-seq & GSE59114 \\
    Mouse & $493 \times 2,406$   & Smart-seq   & scRNA-seq & \\
    Mouse & $158 \times 1,737$   & Fluidigm C1 & scRNA-seq & GSE71982 \\
    Mouse & $117 \times 1,673$   & Fluidigm C1 & scRNA-seq & \\
    Mouse & $85 \times 1,584$    & Fluidigm C1 & scRNA-seq & \\
    Mouse & $85 \times 1,648$    & Fluidigm C1 & scRNA-seq & \\
    
    \midrule
    
    Mouse & $197 \times 3,982$   & Fluidigm C1 & scRNA-seq & GSE74596 \\
    Mouse & $355 \times 3,301$   & Fluidigm C1 & scRNA-seq & GSE67310 \\
    Mouse & $3,694 \times 999$   & Fluidigm C1 & scRNA-seq & GSE75330 \\
    Human & $4,959 \times 1,008$ & Fluidigm C1 & scRNA-seq & \\
    Human & $501 \times 3,523$   & Fluidigm C1 & scRNA-seq & GSE85066 \\
    \midrule
    
    Macaque & $182 \times 5,273$ & SC3-seq & scRNA-seq & GSE74767 \\
    Macaque & $83 \times 4,780$  & SC3-seq & scRNA-seq & \\
    Macaque & $272 \times 5,376$ & SC3-seq & scRNA-seq & \\
    Macaque & $351 \times 5,416$ & SC3-seq & scRNA-seq & \\
             
    \midrule
    
    Mouse & $318 \times 2,702$ & Fluidigm C1 & scRNA-seq & GSE70240, 243, 244, 236 \\
    Mouse & $376 \times 2,967$ & Fluidigm C1 & scRNA-seq & GSE70240, 244, 236 \\
    Mouse & $60 \times 1,849$  & Smart-seq2  & scRNA-seq & GSE79363 \\
    Mouse & $749 \times 1,033$ & Fluidigm C1 & scRNA-seq & GSE67602 \\
    Mouse & $699 \times 1,060$ & Fluidigm C1 & scRNA-seq & \\
    Mouse & $346 \times 1,009$ & Fluidigm C1 & scRNA-seq & \\
    \midrule
    Mouse & $213 \times 4,872$   & Fluidigm C1 & scRNA-seq & GSE90860 \\
    Human & $659 \times 3,630$   & Smart-seq2  & scRNA-seq & GSE86146 \\
    Human & $629 \times 4,691$   & Smart-seq2  & scRNA-seq & \\
    Human & $659 \times 3,630$   & Smart-seq2  & scRNA-seq & \\
    Human & $671 \times 4,459$   & Smart-seq2  & scRNA-seq & \\
    \bottomrule
  \end{tabular}
\end{table*}

\begin{table*}[t]
  \centering
  \renewcommand{\arraystretch}{1.1}
  \setlength{\tabcolsep}{15pt}
  
  \caption{Overview of the single-cell datasets selected from the REAL-GOLD and REAL-SILVER benchmarks (Continued).}
  \label{tab:sc_datasets_part2}
  
  \begin{tabular}{l  c l l r}
    \toprule
    \textbf{Organism} & \textbf{Shape (Cell $\times$ Gene)} & \textbf{Technology} & \textbf{Type} & \textbf{Source} \\

    \midrule

    Mouse & $322 \times 6,763$   & Smart-seq2  & scRNA-seq & GSE87375 \\
    Mouse & $563 \times 6,372$   & Smart-seq2  & scRNA-seq & \\
    Mouse & $503 \times 2,037$   & Smart-seq2  & scRNA-seq & GSE90047 \\
    Mouse & $192 \times 1,534$   & Smart-seq2  & scRNA-seq & GSE99951 \\
    Mouse & $456 \times 3,585$   & Smart-seq2  & scRNA-seq & \\
    
    \midrule
    
    Fly & $277 \times 1,514$     & Smart-seq2  & scRNA-seq & GSE100058 \\
    Fly & $169 \times 982$       & Smart-seq2  & scRNA-seq & \\
    Fly & $454 \times 2,703$     & Smart-seq2  & scRNA-seq & \\
    Mouse & $3,580 \times 698$   & 10X Chromium& scRNA-seq & GSE95315 \\
    Mouse & $414 \times 12,532$  & rdRNA-seq   & scRNA-seq & GSE98664 \\
    
    \bottomrule
  \end{tabular}
\end{table*}
\subsection{Preprocessing Pipeline}
\label{appendix:preprocessing}

To ensure the robustness of downstream trajectory inference tasks and mitigate the technical noise inherent in high-throughput sequencing technologies (e.g., dropouts, amplification bias), we implement a rigorous preprocessing pipeline $\Phi$. Let $\mathcal{D} = \{(\mathbf{x}_i, \mathbf{s}_i)\}_{i=1}^N$ denote the raw dataset, where $\mathbf{x}_i \in \mathbb{R}^M$ represents the gene expression vector for cell $i$ across $M$ genes, and $\mathbf{s}_i \in \mathbb{R}^2$ denotes its spatial coordinates (if applicable). The pipeline transforms this raw input into a denoised, low-dimensional latent representation $\mathbf{Z}$, formally defined as:

\begin{equation}
    \mathbf{Z} = \Phi(\mathbf{X}) = \text{PCA}(\text{SelectFeatures}(\text{LogNorm}(\text{QC}(\mathbf{X})))).
\end{equation}

The detailed protocol typically involves the following four stages:

\subsubsection{Stage 1: Quality Control (QC)}
The first stage filters out low-quality observations that could introduce artifacts into the manifold learning process. We define the set of valid cells $\mathcal{D}'$ by imposing constraints on the library size (total counts) and mitochondrial content. Low library sizes often indicate empty droplets or dead cells, while anomalously high counts suggest doublets. High mitochondrial proportions are a signature of cellular stress or apoptosis.
\begin{equation}
\begin{aligned}
\mathcal{D}' = \{i \mid \quad & \text{min\_genes} < \|\mathbf{x}_i\|_0 < \text{max\_genes} \\
\land \quad & \text{pct\_mt}(\mathbf{x}_i) < \theta_{mt} \}.
\end{aligned}
\end{equation}
where $\|\mathbf{x}_i\|_0$ denotes the number of detected genes (L0 norm), and $\theta_{mt}$ is the threshold for mitochondrial percentage (typically $10\% - 20\%$).

\subsubsection{Stage 2: Normalization and Variance Stabilization}
Since cellular sequencing depths can vary significantly due to technical capture efficiency rather than biological differences, we apply a global scaling normalization. The expression vector for each cell is normalized to a fixed total count $s$:
\begin{equation}
    \mathbf{x}_i' = s \cdot \frac{\mathbf{x}_i}{\sum_j x_{ij}}.
\end{equation}
Subsequently, we apply a natural logarithm transformation to stabilize the variance across the dynamic range of expression values and to transform the data distribution closer to normality:
\begin{equation}
    \mathbf{x}_i'' = \ln(\mathbf{x}_i' + 1).
\end{equation}
The addition of a pseudo-count of 1 preserves sparsity for non-expressed genes.

\subsubsection{Stage 3: Feature Selection (HVGs)}
To reduce computational complexity and focus on biological heterogeneity, we subset the feature space to Highly Variable Genes (HVGs). We model the relationship between mean expression $\mu_j$ and dispersion $\sigma^2_j$ for each gene $j$. Genes with a dispersion significantly higher than expected given their mean expression are retained, forming the subset $G_{HVG}$ (typically $|G_{HVG}| \approx 2000$). This step effectively filters out housekeeping genes (constant expression) and low-abundance noise.

\subsubsection{Stage 4: Latent Manifold Embedding}
Finally, we project the high-dimensional normalized data restricted to $G_{HVG}$ into a lower-dimensional latent space to capture the principal axes of variation. We perform Principal Component Analysis (PCA) via Singular Value Decomposition (SVD) on the centered and scaled data matrix:
\begin{equation}
    \mathbf{Z} = \text{SVD}_k(\mathbf{X}_{HVG} - \bar{\mathbf{X}}_{HVG}).
\end{equation}
The resulting matrix $\mathbf{Z} \in \mathbb{R}^{N \times k}$ (where $k=50$) serves as the compact input representation for the subsequent agent-driven trajectory inference modules. This projection preserves the global topology of the cellular manifold while discarding orthogonal components largely attributed to technical noise.

\subsection{Additional Experiments}
\label{appendix:addexper}
\textbf{LLM-based baselines comparison.} We conducted an additional experiment to compare our framework with the following LLM-based baselines to demonstrate the advantages of our framework:
\begin{itemize}
    \item CellAgent: a LLM-driven multi-agent framework for single-cell RNA sequencing (scRNA-seq) data analysis.
    \item scAgents: a multi-agent framework for fully autonomous end-to-end single-cell analysis.
\end{itemize}
We experiment with five subsets (S1 to S5) from the real gold benchmark and calculate the Spearman rank correlation. The results are in the Table \ref{tab: add_2}. According to the results, our method shows the best performance.

\begin{table}
\centering
\caption{Performance comparison with LLM-based baseline methods.}
\label{tab: add_2}

\begin{tabular}{l c c c c c c}
\toprule
\textbf{Method} & \textbf{S1} & \textbf{S2} & \textbf{S3}& \textbf{S4}& \textbf{S5} \\
\midrule
CellAgent & 0.473& 0.577& 0.577 & 0.481& 0.703 \\
scAgents &0.440 & 0.458 & 0.533 &0.475&0.642\\
\midrule
\textbf{Ours} &\textbf{0.526} & \textbf{0.615} & \textbf{0.571} &\textbf{0.483}&\textbf{0.755} \\
\bottomrule
\end{tabular}%
\end{table}

\textbf{Quantitative biological validation}. To quantitatively validate the biological fidelity of the trajectories results in the main experiment, we performed an enrichment analysis based on our TI results of the three real-world datasets. Specifically, we quantified the GO enrichment p-values for the dynamic driver genes identified along the trajectory. The results are given in the Table \ref{tab: add_1}. The results can statistically prove that the genes discovered by our agent are significantly enriched in the expected biological pathways.

\begin{table}
\centering
\caption{P-values of the GO enrichment analysis on three real-world datasets.}
\label{tab: add_1}

\begin{tabular}{l c c c}
\toprule
\textbf{Metrics } & \textbf{Midbrain} & \textbf{Axolotl} & \textbf{SCI} \\

\midrule
\textbf{P-values} &< 0.001 & < 0.001 & < 0.001 \\
\bottomrule
\end{tabular}%

\end{table}

\subsection{Visualization of Trajectories}
\label{appendix:visualization}
\begin{table*}[t]
  \caption{Definition of the Standardized Trajectory Inference (TI) Workflow. \textmd{This workflow, consisting of six sequential tasks, serves as the benchmark for evaluating Task Success Rate (TSR) in the ablation study.}}
  \label{tab:standard_workflow}
  \centering
  \small 

  \begin{tabular}{l p{0.25\textwidth} p{0.6\textwidth}}
    \toprule
    \textbf{Task ID} & \textbf{Task Name} & \textbf{Description \& Objectives} \\
    \midrule
    \textbf{Task 1} & Data Ingestion and Quality Control & Load raw expression matrix and spatial coordinates. Filter low-quality cells (e.g., high mitochondrial content, low gene counts) to output a cleaned \texttt{AnnData} object. \\
    \addlinespace[0.5em]
    
    \textbf{Task 2} & Preprocessing and Normalization & Normalize data (e.g., CPM) and perform logarithmic transformation. Identify Highly Variable Genes (HVGs) to mitigate feature space noise. \\
    \addlinespace[0.5em]
    
    \textbf{Task 3} & Dimensionality Reduction & Perform Principal Component Analysis (PCA) on HVGs, followed by non-linear reduction (e.g., UMAP, t-SNE) to visualize the global manifold structure. \\
    \addlinespace[0.5em]
    
    \textbf{Task 4} & Clustering and Root Identification & Cluster cells (Leiden/Louvain) and identify the trajectory start point ("root") via automatic marker detection (e.g., stemness genes) or user priors. \\
    \addlinespace[0.5em]
    
    \textbf{Task 5} & Trajectory Inference and Pseudotime Calculation & \textbf{Core Step:} Apply specific TI algorithms (e.g., PAGA, DPT, Monocle) to infer the lineage backbone and calculate pseudotime values for all cells. \\
    \addlinespace[0.5em]
    
    \textbf{Task 6} & Visualization and Result Export & Visualize the inferred trajectory on low-dimensional embeddings (colored by pseudotime) and export the final object for downstream analysis. \\
    \bottomrule
  \end{tabular}
\end{table*}

\begin{figure*}
    \centering
    \includegraphics[width=0.72\linewidth]{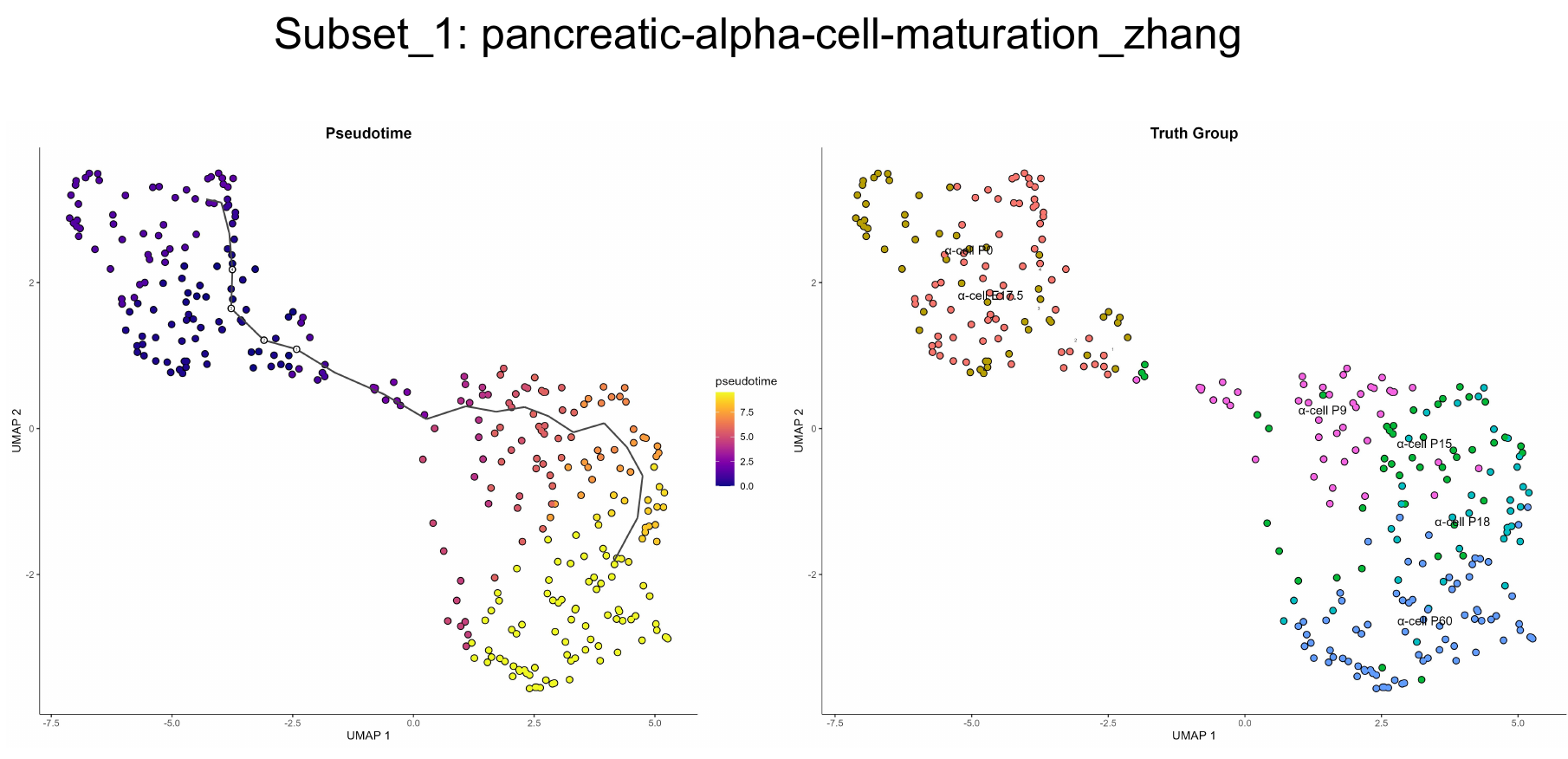}
    \caption{The visualization pseudotime results of TI analysis by SpaCellAgent compared with ground truth on REAL-GOLD dataset.}
    \Description{}
    \label{figapvs5}
\end{figure*}
\begin{figure*}
    \centering
    \includegraphics[width=1\linewidth]{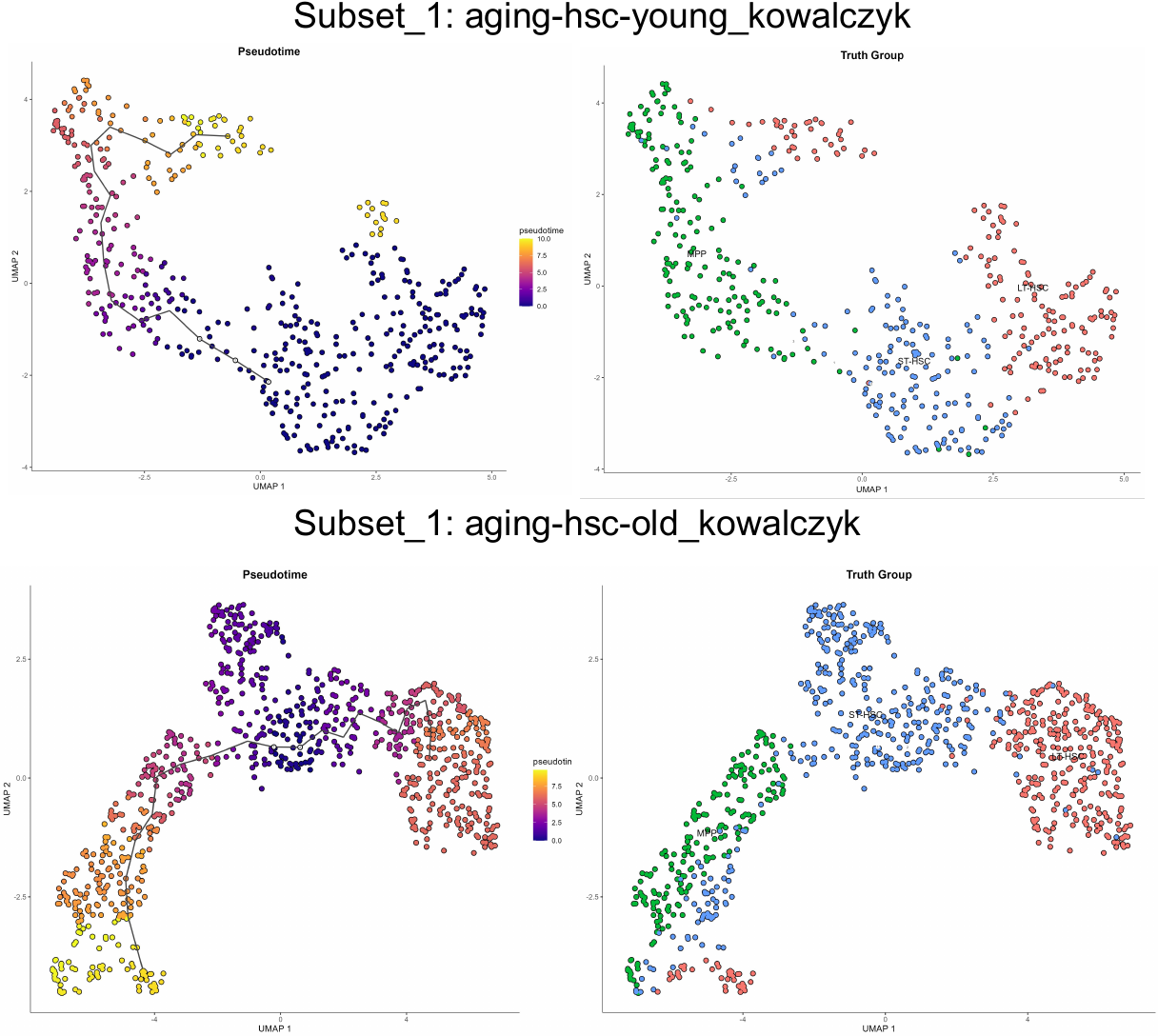}
    \caption{The visualization pseudotime results of TI analysis by SpaCellAgent compared with ground truth on REAL-GOLD dataset.}
    \Description{}
    \label{figapvs4}
\end{figure*}

\begin{figure*}
    \centering
    \includegraphics[width=1\linewidth]{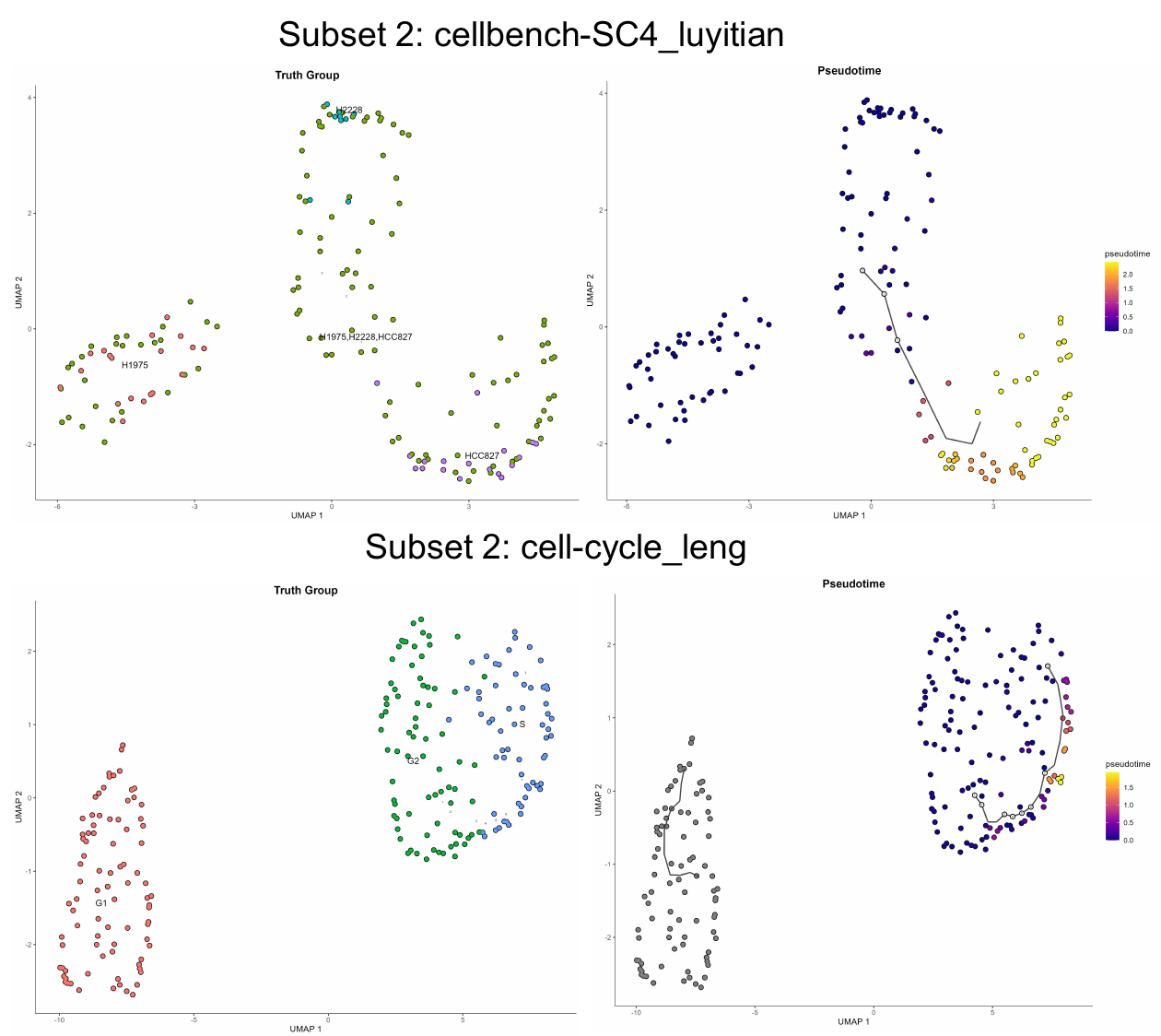}
    \caption{The visualization pseudotime results of TI analysis by SpaCellAgent compared with ground truth on REAL-GOLD dataset.}
    \Description{}
    \label{figapvs1}
\end{figure*}

\begin{figure*}
    \centering
    \includegraphics[width=1\linewidth]{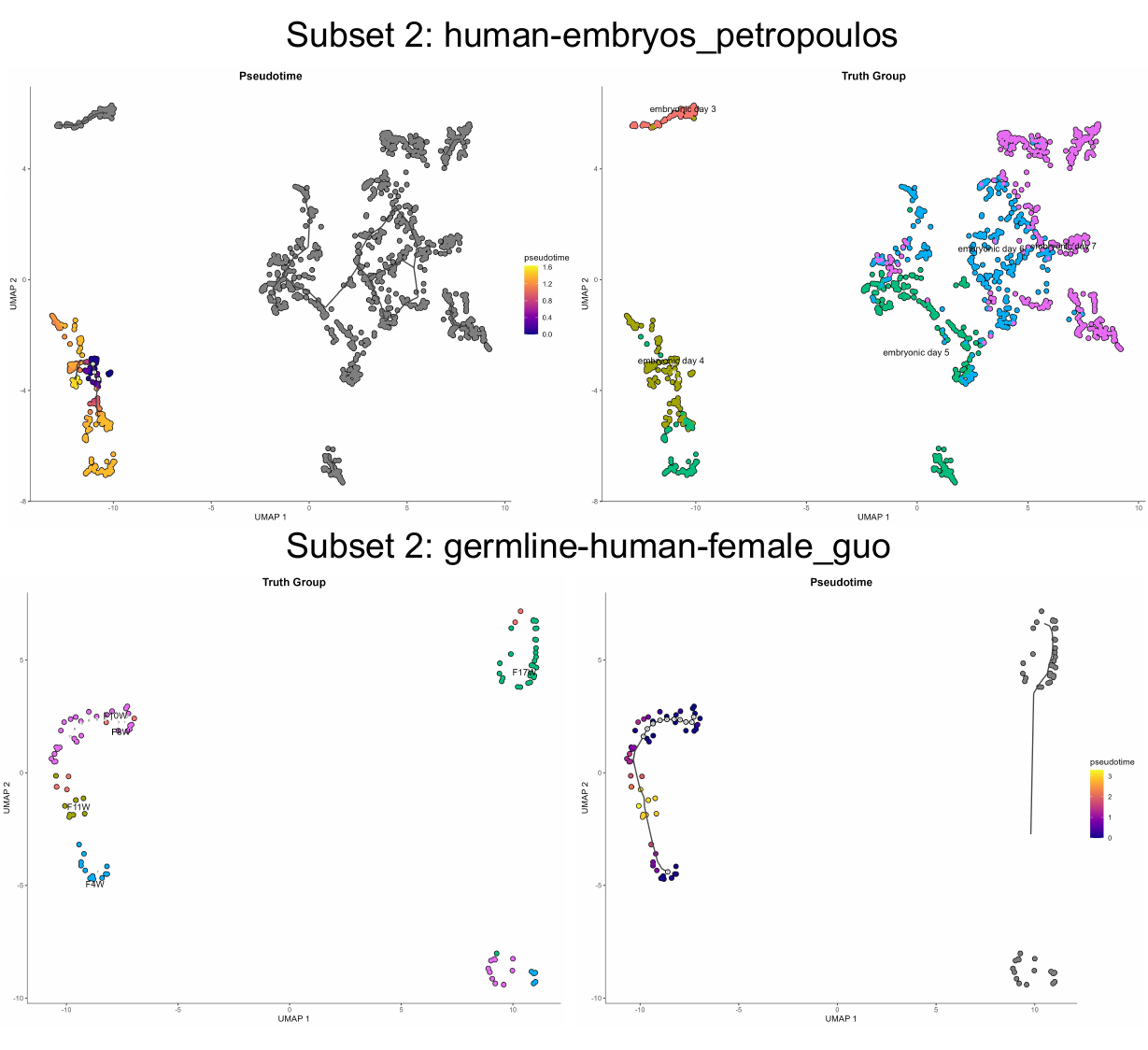}
    \caption{The visualization pseudotime results of TI analysis by SpaCellAgent compared with ground truth on REAL-GOLD dataset.}
    \Description{}
    \label{figapvs2}
\end{figure*}

\begin{figure*}
    \centering
    \includegraphics[width=1\linewidth]{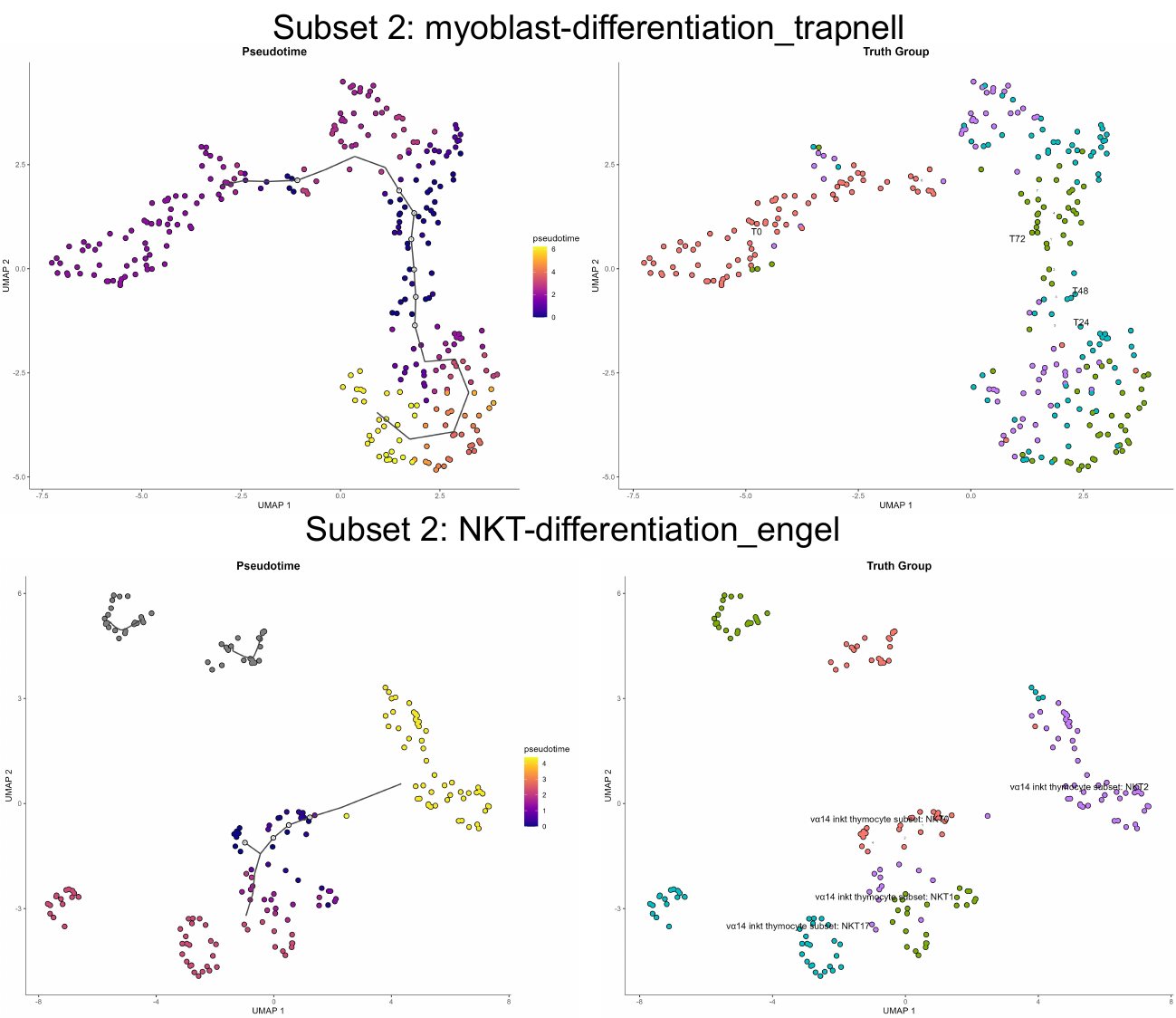}
    \caption{The visualization pseudotime results of TI analysis by SpaCellAgent compared with ground truth on REAL-GOLD dataset.}
    \Description{}
    \label{figapvs3}
\end{figure*}

To provide a comprehensive qualitative assessment of \textit{SpaCellAgent}, we present additional visualizations of the inferred cellular trajectories across benchmarks, as shown in Figure \ref{figapvs1} \ref{figapvs2} \ref{figapvs3} \ref{figapvs4} \ref{figapvs5}. The figures below illustrate the low-dimensional manifold embeddings colored by inferred pseudotime and ground-truth cell types. These visualizations visually substantiate our quantitative findings, demonstrating the framework's capability to correctly recover complex topologies, including linear, bifurcating, and multifurcating lineages, that characterize diverse biological processes.

\subsection{Downstream Analysis}
\begin{figure*}
    \centering
    \includegraphics[width=0.8\linewidth]{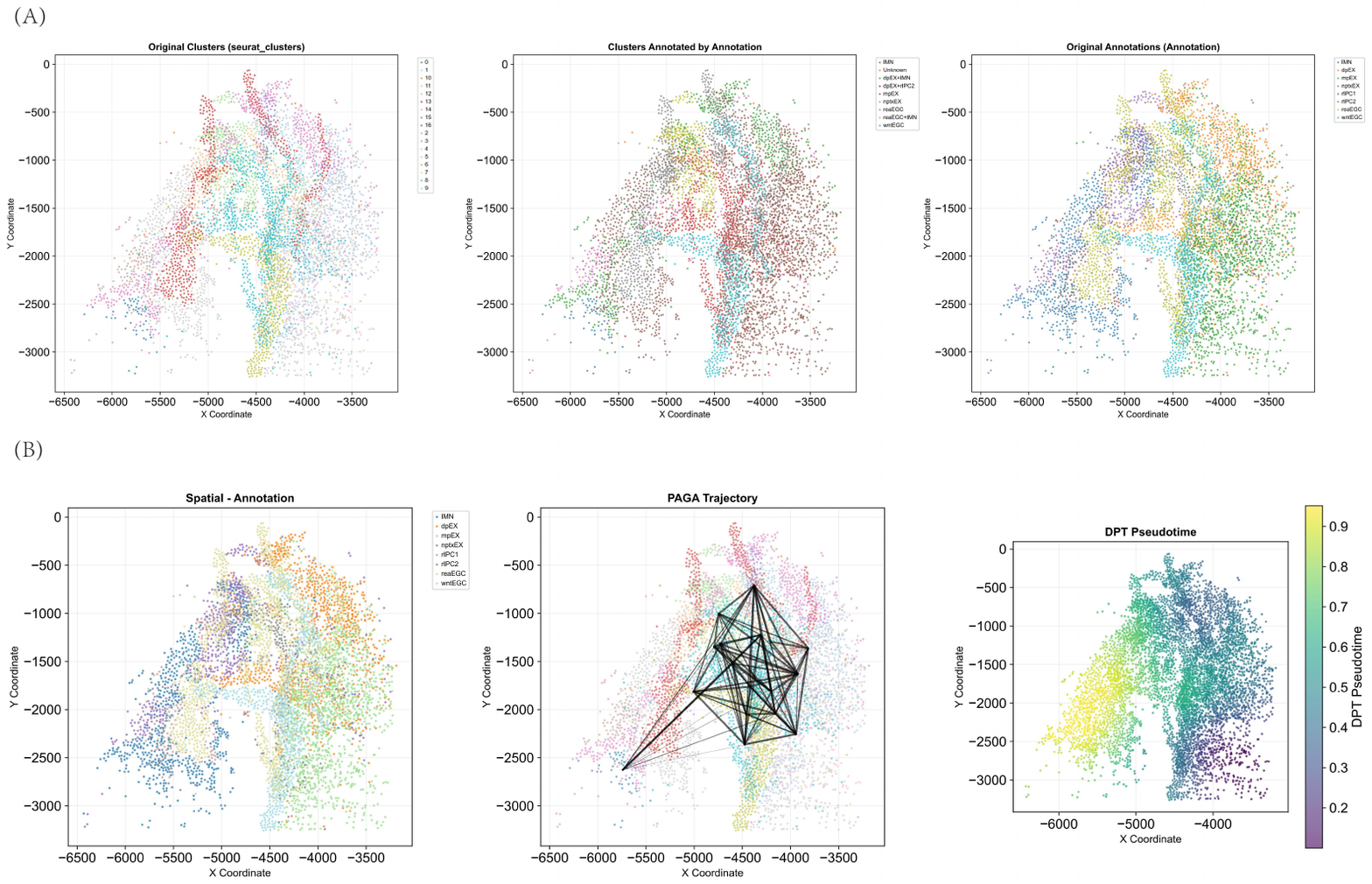}
    \caption{Spatial visualization of clusters, cell annotation by SpaCellAgent and ground truth annotation.}
    \Description{}
    \label{fig:apsp1}
\end{figure*}
\begin{figure*}
    \centering
    \includegraphics[width=0.7\linewidth]{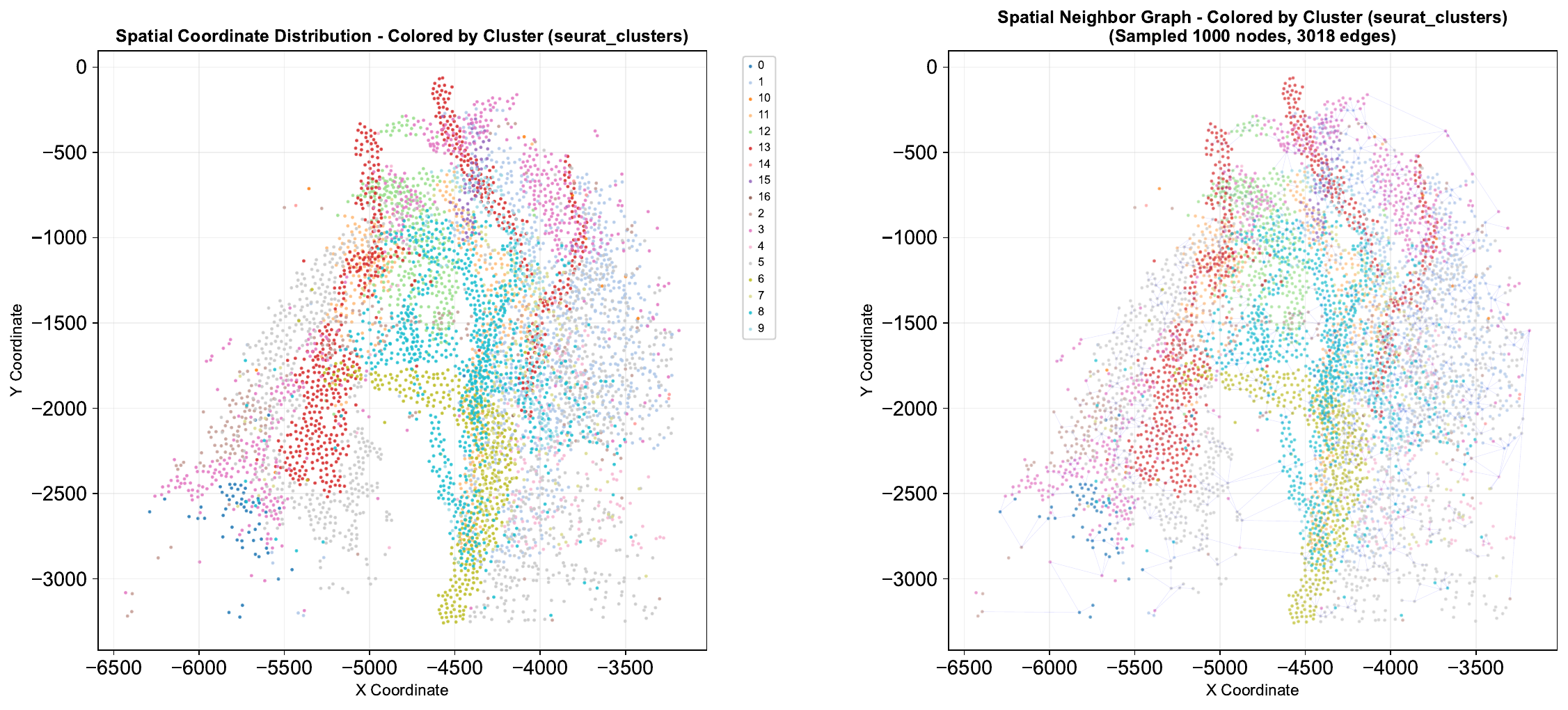}
    \caption{Visualization of spatial topology and neighborhood connectivity from Axolotl Neuron dataset. The left panel displays the spatial coordinate distribution of cells colored by identified clusters, illustrating the physical organization of the tissue. The right panel presents the constructed spatial neighbor graph (visualizing a sampled subgraph of 1,000 nodes and 3,018 edges for clarity), where edges denote physical proximity. This graph structure serves as a critical constraint for downstream analysis, ensuring that inferred trajectories respect the biological reality of spatial continuity.}
    \Description{}
    \label{fig:apsp2}
\end{figure*}

\begin{figure*}
    \centering
    \includegraphics[width=1\linewidth]{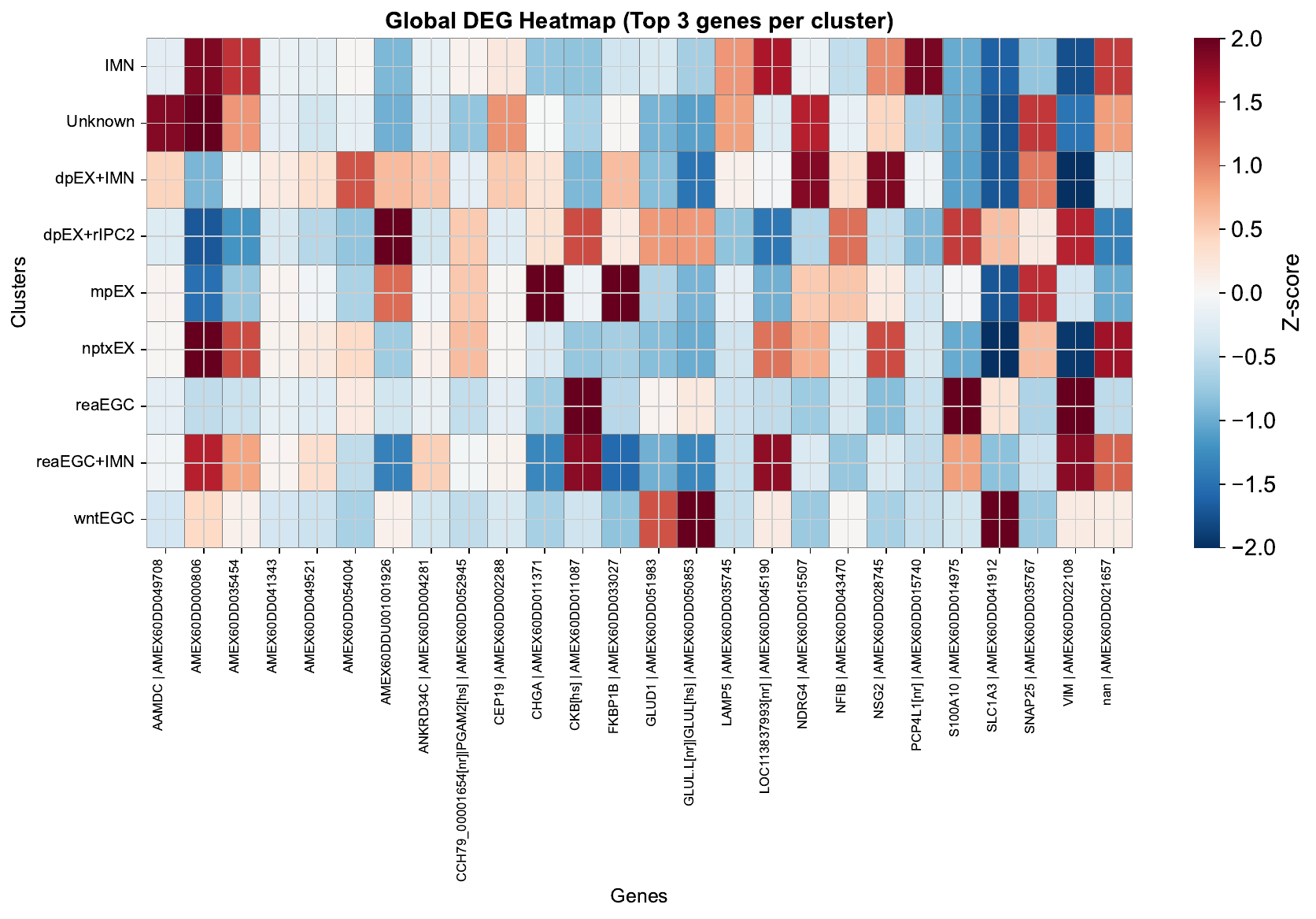}
    \caption{DEG heat map of Axolotl Neuron dataset. The visualization displays the top 3 marker genes identified for each annotated cell type (rows), such as IMN, mEX, and wntEGC. Columns represent specific genes, and the color scale reflects Z-score normalized expression levels, with red indicating upregulation and blue indicating downregulation, highlighting distinct molecular signatures for each cluster.}
    \Description{}
    \label{fig:apsp3}
\end{figure*}
\label{appendix:downstream}
Beyond reconstructing the backbone of cellular development, \textit{SpaCellAgent} autonomously conducts downstream characterization to uncover biological mechanisms. In this section, we include supplementary results of the Axolotl Neuron Regeneration dataset (\ref{fig:apsp1} \ref{fig:apsp2}), and visualizations of gene expression dynamics along the inferred paths of the Mouse Dorsal Midbrain dataset (\ref{fig:apsp3}). The following heatmaps and gene trend plots highlight the agent's ability to identify spatially and temporally variable genes that drive cell fate decisions, confirming the biological plausibility of the reconstructed lineages.
\subsection{Extended results of Case Study}

\begin{figure*}
    \centering
    \includegraphics[width=0.8\linewidth]{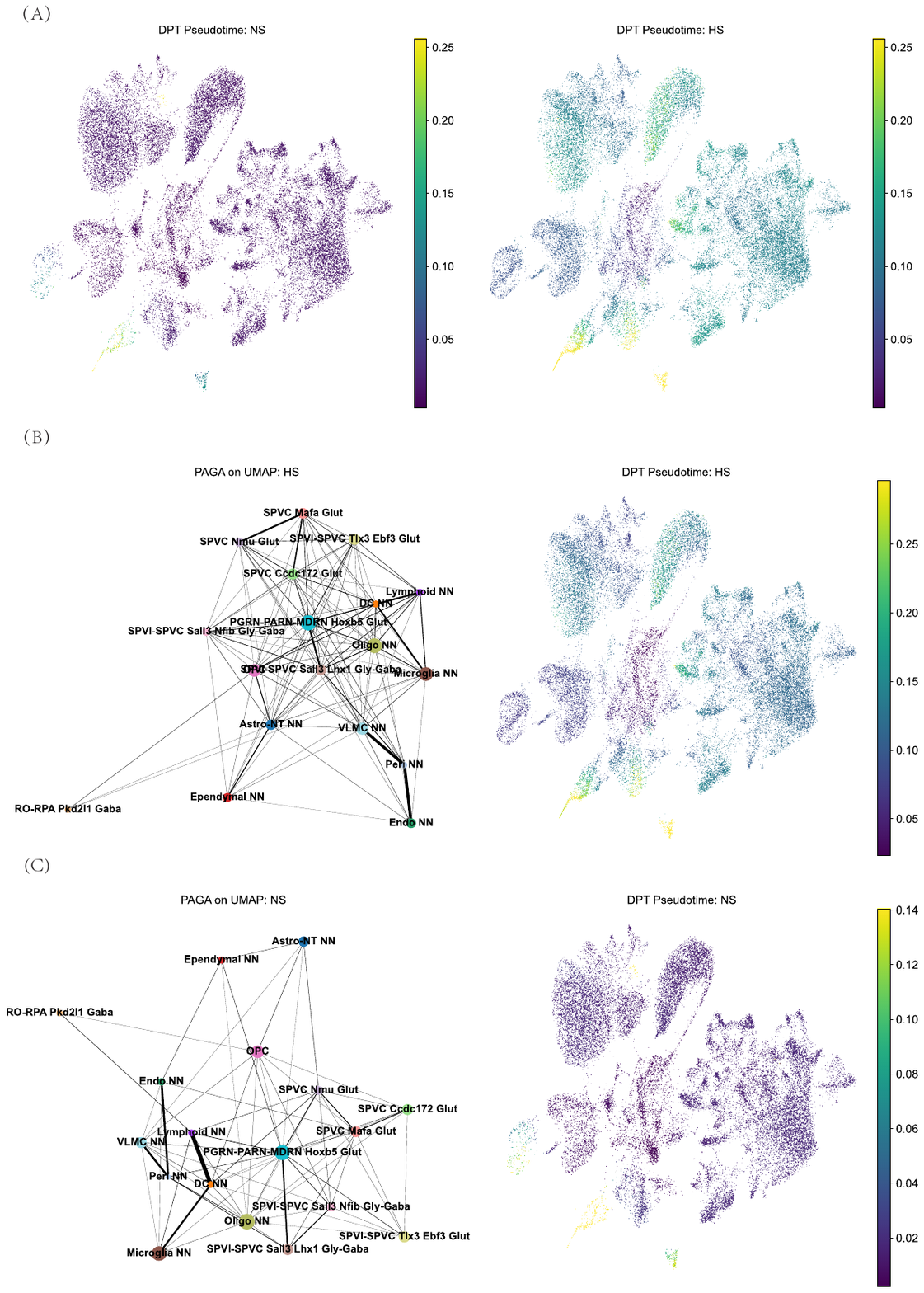}
    \caption{(A) Pseudotime of HS and NS mouse SCI dataset. (B) HS datasets' comparison of TI Umap plot by PAGA and Pseudotime plot. (C) NS datasets' comparison of TI Umap plot by PAGA and Pseudotime plot.}
    \Description{}
    \label{figapca1}
\end{figure*}
\begin{figure*}
    \centering
    \includegraphics[width=0.7\linewidth]{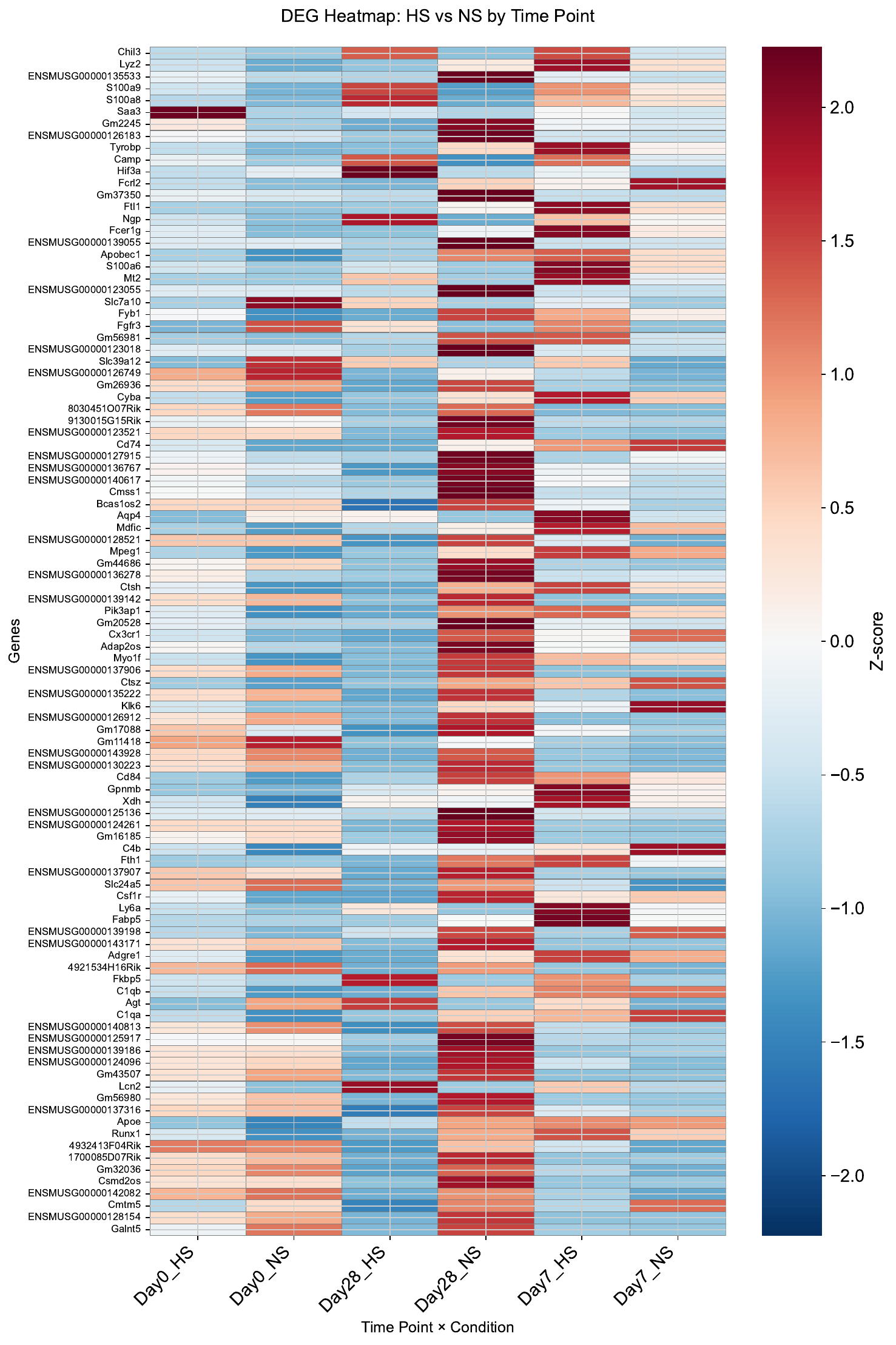}
    \caption{Global DEG heat map of mouse SCI dataset compared with time point and NS/HS.}
    \Description{}
    \label{figapca2}
\end{figure*}
\begin{figure*}
    \centering
    \includegraphics[width=0.7\linewidth]{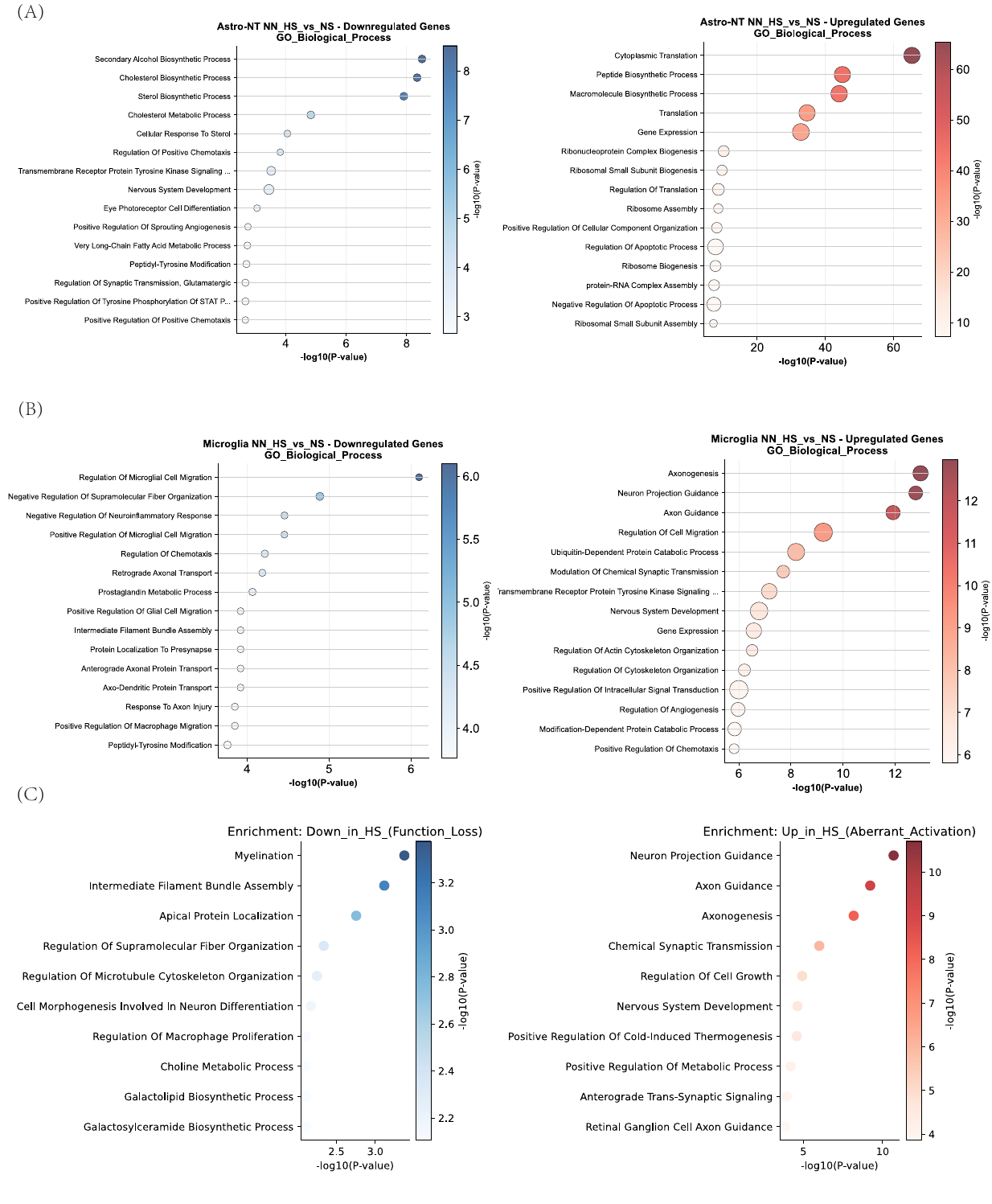}
    \caption{Gene Ontology (GO) enrichment analysis of differentially expressed genes. (A) Enriched biological processes for downregulated (left) and upregulated (right) genes in Astrocytes (Astro-NT). (B) Enriched pathways in Microglia, highlighting the downregulation of regulatory functions and upregulation of axonogenesis-related processes. (C) Functional categorization of key pathways, distinguishing between function loss (e.g., Myelination) and aberrant activation (e.g., Neuron Projection Guidance). The x-axis represents the statistical significance (-log10 P-value).}
    \Description{}
    \label{figapca3}
\end{figure*}
\label{appendix:case}

Due to space constraints in the main manuscript, we provide an extended analysis of the case studies here. We intermediate outputs generated during the analysis of theunpublished Mouse SCI datasets. As shown in Figure \ref{figapca1}, the visualization of the pseudotime plot and PAGA plot shows the global TI of Mouse SCI datasets. Then, we plot the DEG heat map of the whole Mouse SCI dataset using SpaCellAgent, as shown in Figure \ref{figapca2}. Subsequently, we conduct gene ontology enrichment analysis on the whole dataset, as shown in Figure \ref{figapca3}. These results further exemplify the robustness of the self-evolution module, visualizing how the agent detects anomalies, corrects errors, and iteratively refines the trajectory model to align with biological priors.

\subsection{Ablation Studies Details}
\label{abdetail}
\textbf{Expert Selection.}We recruit 5 domain experts: Expert 1 holds a Ph.D. degree in biomedical engineering. Expert 2 to 5 holds Ph.D. degrees in clinical medicine. Crucially, all experts possess over 3 years of hands-on experience in omics data analysis.

\textbf{Standardized Instructions}. Before the experiment, all experts receive a briefing on the task objectives and success criteria: 
\begin{itemize}
    \item Task objectives. They are provided with basic information about the test datasets, a standardized analysis workflow, and the flexibility to select their preferred analytical tools based on personal expertise and preference.
    \item Success criteria. When the expert finishes the TI analysis and validates its biological relevance and significance.
 \end{itemize}
\textbf{Tool Equivalency}. The human experts are equipped with standard SOTA computational tools widely used in the industry. We list these computational tools and provide them to both the human experts and SpaCellAgent.

\textbf{Timing Protocol}. The timer commences exactly when the raw datasets are made available to the user/expert and concludes when the final validated output is submitted. For a rigorous comparison, we partition the standard trajectory inference analysis into five standardized stages: Step 1: Data Preprocessing, Step 2: Batch Correction, Step 3: Cell Type Annotation, Step 4: Dimensionality Reduction and Clustering, Step 5: Trajectory inference and pseudotime calculation.

\textbf{Statistical Variance Analysis}. To strengthen the claim of our 41.2\% improvement, we provide the experiment details of appropriate statistical summaries (mean and standard deviation). The results are in the following table.

To guarantee a fair and rigorous benchmark across all ablation variants (including SpaCellAgent w/o Planner, Evaluator, or Self-Evolution) and the GPT-4 baseline, we established a Standardized TI Workflow. This protocol systematically deconstructs the multifaceted lineage reconstruction process into six distinct, sequential tasks, mirroring the standard bioinformatics pipeline from data ingestion to visualization. Crucially, this granular decomposition allows us to pinpoint specific failure modes, whether they stem from logic errors in pseudotime calculation or syntax errors in plotting, thereby offering a high-resolution view of each component's contribution.
For the quantitative assessment via Task Success Rate (TSR), a variant is deemed successful at a specific step strictly if it generates executable code that yields valid intermediate artifacts without runtime exceptions. The detailed definitions of these six pivotal tasks are enumerated in Table \ref{tab:standard_workflow}.
In our ablation study, the full \textit{SpaCellAgent} autonomously generates this 5-step plan via its planner. In contrast, the \textit{w/o Planner} variant attempts to execute the entire pipeline in a monolithic block, while \textit{GPT-4} operates without the guidance of this structured decomposition.

\begin{table}
  \caption{Definition of the Standardized Trajectory Inference (TI) Workflow. \textmd{This workflow, consisting of five sequential tasks, serves as the benchmark for evaluating Task Success Rate (TSR) in the ablation study.}}
  \label{tab:standard_workflow}
  \centering
  \begin{tabular}{l p{0.25\textwidth} p{0.6\textwidth}}
    \toprule
    \textbf{Task ID} & \textbf{Task Name}  \\
    \midrule
    \textbf{Task 1} & Data Ingestion and Quality Control \\
    \addlinespace[0.5em]
    
    \textbf{Task 2} & Preprocessing and Normalization  \\
    \addlinespace[0.5em]
    
    \textbf{Task 3} & Dimensionality Reduction \\
    \addlinespace[0.5em]
    
    \textbf{Task 4} & Clustering and Root Identification \\
    \addlinespace[0.5em]
    
    \textbf{Task 5} & Trajectory Inference and Pseudotime Calculation  \\

    \bottomrule
  \end{tabular}
\end{table}

\subsection{Additional Ablation Studies}
\label{appendix_addab}
To quantitatively verify the effectiveness of the Knowledge-augmented fallback mechanism, we conduct an ablation study with the w/o experience memory variants on the real-gold dataset. We calculate the Spearman's rank correlation, p-value, and the total time cost. The results are shown in Table \ref{tab: add_ab_1}.

\begin{table}
\centering
\caption{ablation study with the w/o experience memory variants.}
\label{tab: add_ab_1}
\resizebox{\columnwidth}{!}{%
\begin{tabular}{l c c c}
\toprule
\textbf{Model } & \textbf{correlation} & \textbf{P-value} & \textbf{Time cost} \\

\midrule
w/o experience memory &0.526 & < 0.001 & 46.0 \\
\midrule
\textbf{Ours} &\textbf{0.538} & \textbf{< 0.001} & \textbf{38.0}\\
\bottomrule
\end{tabular}%
}
\end{table}

\section{Additional Related Works}
\textbf{Recent Advancements in Robust Learning and Reasoning.}
In recent years, significant progress has been made in enhancing machine learning performance under complex data constraints and reasoning tasks. Addressing the challenges of data scarcity and distribution shifts, researchers have developed innovative frameworks for few-shot hypothesis adaptation \cite{chi2021tohan, yang2025nt} and novel class discovery \cite{chimeta, chi2024does}, which effectively improve model generalization in open-world scenarios. Beyond statistical learning, there is a growing trend towards integrating causal mechanisms into deep learning. For instance, Li et al. \cite{litransformer} introduced transformer-based architectures for spatial-temporal counterfactual estimation, while Chi et al. \cite{chi2024unveiling} conducted comprehensive evaluations to unveil the underlying causal reasoning capabilities of large language models.

\end{document}